\documentclass[letterpaper,twocolumn,10pt]{article}
\usepackage{usenix}
\usepackage{amsmath}
\usepackage{amssymb}
\usepackage{algorithmic}
\usepackage{todonotes}
\usepackage{textcomp}
\usepackage{fancyhdr}
\usepackage[bottom]{footmisc}
\usepackage[normalem]{ulem}
\usepackage{scalerel}
\usepackage{upgreek}
\usepackage{footnote}
\usepackage{enumitem}
\usepackage{graphicx}
\usepackage{xcolor}
\usepackage{xspace}
\usepackage{array}
\usepackage{hhline}
\usepackage{multicol}
\usepackage{multirow}
\usepackage{courier}
\usepackage{tikz} 
\usepackage{booktabs}
\usepackage{longfbox}
\usepackage{colortbl}
\usepackage{soul}
\usepackage{comment}
\usepackage{arydshln}
\usepackage{subfig}
\usepackage{adjustbox}
\usepackage{kotex}
\usepackage{setspace}
\usepackage{balance}

\DeclareGraphicsExtensions{.pdf,.png,.jpg}

\newcommand\prop{DeepSketch\xspace}
\newcommand{\usec}{$\mu$s\xspace} 
\newcommand{\cluster}{DK-Clustering\xspace}
\newcommand\pc[1]{\textsf{PC#1}\xspace}
\newcommand\inst{\textsf{Install}\xspace}
\newcommand\upd{\textsf{Update}\xspace}
\newcommand\synth{\textsf{Synth}\xspace}
\newcommand\sensor{\textsf{Sensor}\xspace}
\newcommand\web{\textsf{Web}\xspace}
\newcommand\sof[1]{\textsf{SOF#1}\xspace}
\newcommand\tblk{$T_\text{BLK}$\xspace}

\newcommand{\sect}[1]{{Section~#1}\xspace} 
\newcommand{\fig}[1]{{Figure~#1}\xspace} 
\newcommand{\tab}[1]{{Table~#1}\xspace} 
\newcommand{\head}[1]{{\noindent\textbf{#1.}\xspace}} 

\newcommand{\js}[1]{{\color{black}{#1}}}

\newcommand{\sj}[1]{{\color{black}{#1}}}
\newcommand{\sjout}[1]{{\color{black}{}}}
\newcommand{\om}[1]{{\color{black}{#1}}}
\newcommand{\fix}[1]{{\color{black}{#1}}}

\newcommand{\rev}[1]{{\color{black}{#1}}}

\newcommand{\eg}{\textit{e.g.,}}

\newcommand{\ie}{\textit{i.e.,}}

\newcommand\affileth{$^1$}
\newcommand\affildgist{$^2$}

\newcolumntype{?}{!{\vrule width 1pt}}
\newcolumntype{;}{!{\vrule width 0.5pt}}

\DeclareRobustCommand\bcirc[1]{\tikz[baseline=(char.base)]{
           \node[shape=circle,draw,inner sep=0pt,fill=black, text=white] (char) {#1};}}
\DeclareRobustCommand\rcirc[1]{\tikz[baseline=(char.base)]{
           \node[shape=circle, inner sep=0pt,fill=red, text=white] (char) {#1};}}
\DeclareRobustCommand\blcirc[1]{\tikz[baseline=(char.base)]{
           \node[shape=circle, inner sep=0pt,fill=blue, text=white] (char) {#1};}}

\begin{document}

\date{}

\title{\vspace{-1em} \Large \bf \prop: A New Machine Learning-Based Reference Search Technique\\for Post-Deduplication Delta Compression \vspace{-1.5em}}

\author{
{\rm Jisung Park\affileth\thanks{J. Park and J. Kim are co-primary authors.} \quad Jeonggyun Kim\affildgist\om{\footnotemark[1]} \quad Yeseong Kim\affildgist \quad Sungjin Lee\affildgist \quad Onur Mutlu\affileth}\\
{\affileth{}ETH Z\"urich \quad\quad \affildgist{}DGIST}
}


\maketitle

\setstretch{0.96}

\begin{abstract}
Data reduction in storage systems is becoming increasingly important as an effective solution to minimize the management cost of a data center. 
To maximize data-reduction efficiency, existing post-deduplication delta-compression techniques perform delta compression along with traditional data deduplication and lossless compression.
Unfortunately, we observe that existing techniques achieve significantly lower data-reduction ratios than the optimal due to their limited accuracy in identifying similar data blocks.

In this paper, we propose \emph{\prop}, a new reference search technique for post-deduplication delta compression that leverages the learning-to-hash method to achieve higher accuracy in reference search for delta compression, thereby improving data-reduction efficiency. 
\prop uses a deep neural network to extract a data block's \emph{sketch}, \ie~to create an approximate data signature of the block that can preserve similarity with other blocks. 
Our evaluation using eleven real-world workloads shows that \prop improves the data-reduction ratio by up to 33\% (21\% on average) over a state-of-the-art post-deduplication delta-compression technique.
\end{abstract}

\maketitle
\pagestyle{plain}
\thispagestyle{empty}

\vspace{-1em}
\section{Introduction}\label{sec:introduction}
\vspace{-.5em}
As modern data centers generate a tremendous volume of new data every day, it becomes critical for sustainability to store such large amounts of data in an economical way.
Employing a data-reduction technique is one of the effective solutions to cut down the management cost of a data center.
A data-reduction technique reduces the amount of data physically stored in storage media by reducing data redundancy, which allows a data center to handle the same amount of data with fewer or smaller resources (\eg~storage devices and servers).

Many prior works have proposed various data-reduction techniques based on data compression~\cite{lee-ieeetce-2011, liu-ieeetce-2018, lin-fast-2014, horn-nsdi-2017, makatos-eurosys-2010, burrows-asplos-1992} and data deduplication~\cite{quinlan-fast-2002, jain-fast-2005, zhu-fast-2008, dubnicki-fast-2009, lillibridge-fast-2009, dong-fast-2011, meyer-fast-2011, chen-fast-2011, gupta-fast-2011, srinivasan-fast-2012, mandal-fast-2016}.
Data compression encodes a data block using lossless-compression algorithms so that a smaller number of bits can represent the data block.  
Data deduplication prevents a data block from being written if there already exists an \emph{identical} data block (\ie~a block that contains exactly the same data) in the storage system.
To achieve a high data-reduction ratio (\ie~\emph{Original~Data~Size~/~Reduced~Data~Size}), some studies~\cite{lee-date-2013, ma-trustcom-2016} integrate the two techniques in a manner that first applies data deduplication for incoming (\ie~to-be-stored) blocks and performs lossless compression on non-deduplicated blocks.

Delta compression~\cite{wu-eurosys-2012, ajtai-jacm-2002, burns-ieeetkde-2003, park-date-2017, shilane-fast-2012, xia-ieeetc-2015, zhang-fast-2019} has recently received increasing attention as a complementary method to overcome the limitations of data compression and data deduplication.  
It compares the data block to compress with a \emph{reference} data block and extracts only different bit patterns between the two blocks, which are then encoded using lossless compression.  
The more similar the data block and the reference (\ie~the smaller the delta between the data and the reference), the higher the data-reduction ratio.  
By leveraging the similarity between two blocks, delta compression can achieve a high data-reduction ratio even for non-duplicate data (which cannot benefit from data deduplication) and high-entropy data (which lossless compression cannot efficiently handle).  
Several prior works~\cite{park-date-2017, shilane-fast-2012, xia-ieeetc-2015, zhang-fast-2019} demonstrate that \textit{post-deduplication delta compression}, which performs deduplication, delta compression, and lossless compression in order, can significantly improve the data-reduction ratio over simple integration of deduplication and lossless compression.

A key challenge for post-deduplication delta-compression techniques is how to find a good reference block that provides a high data-reduction ratio.
The most intuitive approach is to scan all the data blocks stored in the storage system and use the one that provides the highest data-reduction ratio as the reference for the incoming block.
Unfortunately, doing so is practically infeasible due to its prohibitive performance overhead. 
To address this, prior works~\cite{park-date-2017, shilane-fast-2012, xia-ieeetc-2015, zhang-fast-2019} use \emph{locality-sensitive hash (LSH)} functions~\cite{indyk-stoc-1997, broder-jcss-2000} to generate similar data signatures for data blocks with similar bit patterns, which is called \emph{data sketching}.
Data sketching enables quick reference search across a large-scale storage system by comparing only the signatures (\ie~sketches) of data blocks.

In this work, we observe that existing post-deduplication delta-compression techniques~\cite{zhang-fast-2019, shilane-fast-2012} achieve significantly lower data-reduction ratios than the optimal due to the high \emph{false-negative rate (FNR)} of LSH-based reference search.
Our analysis using six real-world workloads shows that, although a state-of-the-art reference search technique~\cite{zhang-fast-2019} is effective at identifying a \emph{very similar} reference block (which thus provides a very high data-reduction ratio) for an incoming block, it also fails to find \emph{any} reference block for a large number of incoming blocks (up to 75.5\%) that actually have at least one good reference block within the storage system.
Tuning the used LSH function may be able to reduce the high FNR in reference search, but it would require significant human \om{effort} to identify the best settings for \om{each workload}.

\textbf{Our goal} is to improve the space efficiency of a storage system by increasing the accuracy of reference search in post-deduplcation delta compression, thereby reducing the gap between existing data-reduction techniques and the optimal.\footnote{
In this work, we focus on data reduction rather than other optimizations (\eg~mitigation of performance/memory overheads), targeting systems where \om{space} efficiency is the highest priority (\eg~archival or backup systems).}
To this end, we propose \emph{\prop}, a new machine learning (ML)-based data sketching mechanism specialized for reference search in delta compression.\linebreak
\textbf{Our key idea} is to use the \emph{learning-to-hash} method~\cite{kulis-neurips-2009, wang-ieeetpami-2017} to automatically generate similar data signatures for any two data blocks that would provide a high data-reduction ratio when delta-compressed relative to each other.

For each incoming data block, \prop generates the block's sketch using a deep neural network (DNN) model. 
It performs DNN inference with the target data block as an input of the DNN and uses the resulting activation values \om{in} the DNN's last hidden layer as the data block's sketch.
We envision that \prop's DNN is \emph{pre-trained} before building or updating a \prop-enabled system, using data sets collected from other existing systems that store similar (or the same) types of data.

While many prior works~\cite{zhu-AAAI-2016, cao-AAAI-2017, lin-cvpr-2015, cao-cvpr-2018, ryali2020bio, liang-atc-2019} demonstrate the high effectiveness of the learning-to-hash method in various nearest-neighbor search applications (\eg~image recognition and classification), applying the learning-to-hash method to the reference search problem in post-deduplication delta compression is not straightforward.
A key problem is that, unlike existing ML-based applications that deal with specific known data types (\eg~images and voices), \prop needs to process \emph{general binary data}, which introduces two key challenges. 
First, there is no well-defined labeled data or semantic information (\eg~cats, dogs, and monkeys in image classification) within our target data sets. 
Second, possible bit patterns of a data block have an extremely high dimensional space (\eg~$2^{4,096\times8}$ unique bit patterns for a 4-KiB data block). 
Due to the high dimensionality of the target data set, it is difficult to collect large enough data to train the DNN with high inference accuracy using known training methods.

To address the above challenges, we develop a new method to train the DNN of \prop, which generates hash values suitable for reference search in post-deduplication delta compression.
We extend the traditional unsupervised learning approach~\cite{hinton-MIT-1999} in three \om{ways}.
First, based on the k-means clustering algorithm~\cite{lloyd-ieeetit-1982}, we design a new clustering method, called \emph{dynamic k-means clustering (DK-Clustering)}, which effectively classifies high-dimensional data without any knowledge of the number of clusters.
Second, after clustering, we \om{ensure} each cluster to have a sufficient number of data blocks by adding data blocks slightly and randomly modified from each cluster's representative block.
Doing so prevents DNN training from being biased towards some specific data patterns that occur frequently.
Third, we perform two-stage DNN training to enable \prop to generate a data block's hash value. 
We first train a DNN to classify data blocks into the clusters formed by DK-Clustering and then transfer the knowledge of the trained DNN to build the learning-to-hash network that generates the hash values (\ie~sketches) of data blocks.

We integrate our \prop engine into a state-of-the-art post-dedupl\-ication delta-compre\-ssion technique~\cite{zhang-fast-2019}.
Unlike existing techniques that aim to find a reference block whose sketch \emph{exactly} matches that of the incoming block, we exploit a state-of-the-art \emph{approximate} nearest-neighbor search algorithm~\cite{ngt}.
Doing so allows \prop to tolerate small differences within data sketches \rev{(i.e., it can identify similar blocks even when the blocks' sketches are different)}, thereby increasing the chance of delta compression for an incoming data block.
Our evaluation using \fix{eleven} real-world workloads shows that \prop improves the data-reduction ratio by up to 33\% (21\% on average) over the state-of-the-art baseline.

This paper makes the following key contributions:
\begin{itemize}[leftmargin=*, noitemsep, topsep=0pt]
    \item We propose \prop, the first machine learning-based reference search technique for post-deduplication delta compression.
    We demonstrate that the learning-to-hash method can be an effective solution to generate delta-compression-aware data signatures for general binary data.
    \item We introduce a new training method that allows \prop to learn delta-compression-aware data representation for an extremely high-dimensional data set.
    \item We integrate \prop into the state-of-the-art post-dedupli\-cation delta-compression technique~\cite{zhang-fast-2019}.
    Evaluation results using \fix{eleven} real-world workloads show that \prop improves the data-reduction ratio by up to 33\% (21\% on average) compared to the state-of-the-art baseline.
\end{itemize}

\vspace{-1em}
\section{Background\label{sec:bg}}
\vspace{-.5em}
We provide brief background on data-reduction techniques in storage systems necessary to understand the rest of the paper.

\vspace{-1em}
\subsection{Data Reduction in Storage Systems\label{subsec:bg_indv}}
\vspace{-.5em}
There are three major data-reduction approaches: 1) data deduplication, 2) lossless compression, and 3) delta compression.

\head{Data Deduplication}
Data deduplication\cite{quinlan-fast-2002, jain-fast-2005, zhu-fast-2008, dubnicki-fast-2009, lillibridge-fast-2009, dong-fast-2011, meyer-fast-2011, chen-fast-2011, gupta-fast-2011, srinivasan-fast-2012, mandal-fast-2016} reduces the amount of data physically written to storage devices by eliminating duplicate data in the storage system.  
In data deduplication, an incoming data block is not physically written if it has the same data content as a data block previously stored in the storage system. 
Instead, the storage system maintains a table that stores mapping information between such a deduplicated block and the previously-stored block with the same content (called \emph{reference}), so that future reads to any deduplicated blocks can be serviced from their reference.
This mechanism allows data deduplication to store only a single copy of any \om{block-granularity} unique data content in the storage system.

To quickly identify an incoming block's reference, deduplication uses a strong hash function (\eg~SHA1~\cite{shs-fips-1992} or MD5~\cite{rivest-1992}) to generate a data block's \emph{unique} signature, commonly called a~\emph{fingerprint}.
Given two blocks, deduplication determines whether or not they have the same \om{content}, by comparing only the two blocks' fingerprints.
To avoid any data loss due to hash collision, it is common practice for deduplication to use a strong hash function to generate fingerprints whose collision rate is lower than the uncorrectable bit-error rate (UBER) requirement of a disk (\eg~$<10^{-15}$ to $10^{-16}$~\cite{quinlan-fast-2002, gupta-fast-2011, gray-arxiv-2007, cox-fms-2018}).

\head{Lossless Compression}
Data compression~\cite{ziv-ieeetit-1978, huffman-procire-1952, shannon-bell-1948} is a fundamental method to reduce the size of data in computing systems. 
Given a data block, it encodes the block's content to be represented by a smaller number of bits in a manner that replaces repetitive bit patterns with \om{smaller} metadata or symbols. 
Doing so results in an increase in the \emph{entropy}~\cite{shannon-bell-1948} of the compressed data. 
For a data block with low entropy (\ie~the block contains many repeated bit patterns), lossless compression can achieve a high \emph{data-reduction ratio} (\ie~\emph{Original Data Size / Compressed Data Size}).


\head{Delta Compression}
Delta compression~\cite{wu-eurosys-2012, ajtai-jacm-2002, burns-ieeetkde-2003, park-date-2017, shilane-fast-2012, xia-ieeetc-2015, zhang-fast-2019}
eliminates redundant bit patterns that coexist in two different blocks.
It stores only either of the two blocks and the difference (\ie~\emph{delta}) between the two blocks.
Leveraging the \emph{similarity} of two different data blocks enables delta compression to achieve higher data reduction over 1) deduplication, which removes only \emph{identical} data blocks, and 2) lossless compression, which eliminates redundancy only \emph{within a block} and does \emph{not} work well with high-entropy data.
For this reason, delta compression has gained increasing attention in recent studies~\cite{park-date-2017, shilane-fast-2012, xia-ieeetc-2015, zhang-fast-2019} as a complementary method bridging deduplication and lossless compression.

A key challenge for delta compression in large-capacity storage systems is how to find a good reference block that provides a high data-reduction ratio for each incoming block.
Designing a reference search technique for delta compression is similar to solving a nearest-neighbor search problem, as its goal is to find the most similar data block (which does not have to be an \emph{exact} match) within a large data set for a given incoming block.  
The most widely-used approach is to use locality sensitive hashing (LSH)~\cite{indyk-stoc-1997, broder-jcss-2000} to generate a \emph{sketch} of a block~\cite{park-date-2017, shilane-fast-2012, xia-ieeetc-2015, zhang-fast-2019}, which is a \emph{more approximate} signature than the block's fingerprint (used in deduplication for exact-match searching).  
An LSH function $L(d_i)$ is designed to hash data $d_i$, such that the more similar the given data $d_1$ and $d_2$, the lower the bit-pattern difference between $L(d_1)$ and $L(d_2)$.  
LSH-based data sketching enables quick reference search by comparing only the sketches of data blocks.

\vspace{-1em}
\subsection{Combined Data-Reduction Technique\label{subsec:bg_integrated}}
\begin{figure}[t]
    \centering
    \includegraphics[width=1\linewidth]{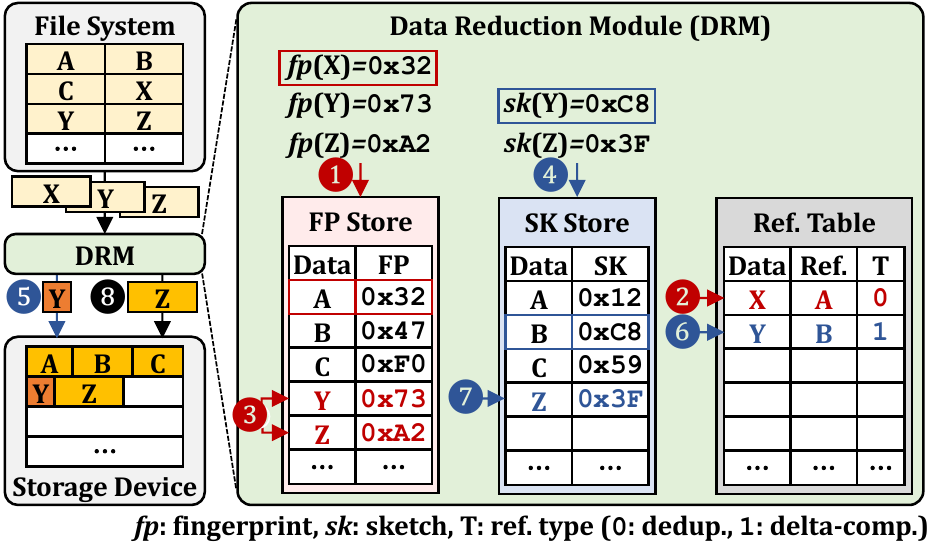}
    \vspace{-2em}
    \caption{Overview of post-deduplication delta compression.}
    \vspace{-2em}
    \label{fig:baseline}
\end{figure}

\vspace{-.5em}
For systems where storage efficiency is the paramount requirement, prior works propose to combine the three major data-reduction techniques, called \emph{post-deduplication delta compression}.
\fig{\ref{fig:baseline}} depicts the overall architecture of a storage system that adopts \om{post-deduplication delta compression}~\cite{shilane-fast-2012, zhang-fast-2019} to perform deduplication (\rcirc{1}--\rcirc{3}), delta compression (\blcirc{4}--\blcirc{7}), and lossless compression (\bcirc{8}) in order.
A data-reduction module (DRM), which can be implemented as an intermediate layer
between a file system and storage devices, performs post-deduplication delta compression for a write request to reduce its size.  
For a read request, it looks for the location of the corresponding compressed data in storage devices and returns the decompressed data.
To this end, the DRM maintains a fingerprint (FP) store and a sketch (SK) store for quick reference search for deduplication and delta compression, respectively, along with a reference (Ref.) table to store the mapping information between each deduplicated or delta-compressed block and its reference block.

For each incoming data block, the DRM \rcirc{1} first checks if the storage system already contains a data block with the same content by referring to the FP store.
If the incoming block's fingerprint matches one in the FP store (\eg~block X matching block A in \fig{\ref{fig:baseline}}), the DRM skips writing the block to the storage device and \rcirc{2} just updates its mapping information in the reference table in order to redirect future reads for the incoming block to the matching reference block.
To use non-deduplicated blocks (\eg~blocks Y and Z) as potential reference data for deduplication in the future, the DRM \rcirc{3} writes their fingerprints into the FP store.

If there is no matching fingerprint in the FP store, the DRM \blcirc{4} searches for matching sketches in the SK store to find a reference block for delta compression.
When it finds a reference block (\eg~block B for block Y), the DRM \blcirc{5} performs delta compression with the reference and stores the compressed data.
There is a possibility of having multiple matching references in the SK store (see Section~\ref{subsec:motiv_lsh}).
In such a case, the DRM usually selects the first-found candidate (called \emph{first-fit} selection)~\cite{zhang-fast-2019, shilane-fast-2012}.
The DRM then \blcirc{6} updates the reference table to map the incoming block to the reference block so that it can decompress the delta-compressed data using the reference block for future read requests.
If no matching sketch is found in the SK store (\eg~block Z), the DRM \blcirc{7} adds the incoming block's sketch into the SK store to use the incoming block as a potential reference block for delta compression in the future.
Finally, the DRM \bcirc{8} compresses the block with a lossless compression algorithm and stores the result.  

\vspace{-1em}
\section{Motivation\label{sec:motiv}}
\vspace{-.5em}
In this section, we discuss 1) the limitations of existing LSH-based post-deduplication delta-compression techniques~\cite{shilane-fast-2012, zhang-fast-2019} and 2) the potential of the learning-to-hash method~\cite{kulis-neurips-2009, wang-ieeetpami-2017} for more accurate reference search in delta compression.

\vspace{-1em}
\subsection{Limitations of LSH-Based Sketching\label{subsec:motiv_lsh}}
\begin{figure}[!t]
    \centering
    \includegraphics[width=\linewidth]{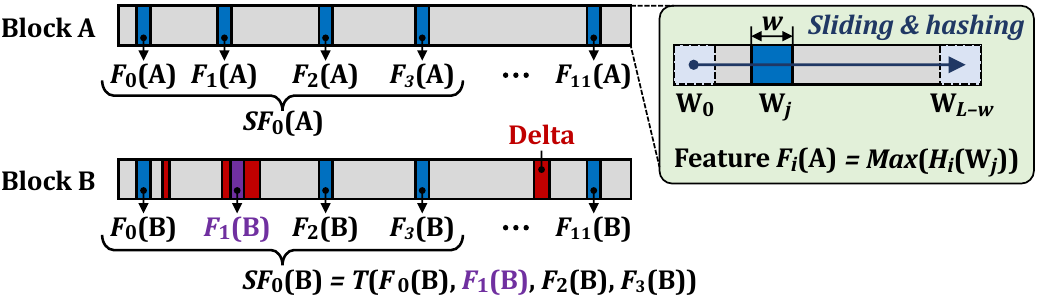}
    \vspace{-2em}
    \caption{An example of LSH-based sketching.}
    \vspace{-1.5em}
    \label{fig:lsh}
\end{figure}

\vspace{-.5em}
As explained in \sect{\ref{subsec:bg_indv}}, LSH-based data sketching enables quick search for a reference block (\ie~reference search) in post-deduplication delta compression.  
\fig{\ref{fig:lsh}} describes the high-level idea of state-of-the-art LSH-based sketching schemes~\cite{zhang-fast-2019, shilane-fast-2012}, which we call \emph{super-feature data sketching (SFSketch)}.
SFSketch generates a data block's sketch using $m$ \emph{features} extracted from the block (\eg~$m=12$ in \fig{\ref{fig:lsh}}).
To extract a feature $F_{i}$(A) of block~A ($ 0 \leq i \leq m - 1$), as shown in \fig{\ref{fig:lsh}} (right), SFSketch calculates the hash value $H_{i}$(W$_{j}$) of each sliding window W$_j$, where $j$ is the starting byte position of the window in the block.
Given a block size of $L$ and a window size of $w$, ($L-w+1$) hash values are calculated in total, and the maximum hash value $Max(H_i(W_j))$ is selected as feature $F_i$(A). 
SFSketch repeats this process to extract $m$ features using a different hash function for each feature (\ie~$H_i$ for $F_i$) and then builds $N$ super-features (SFs) by transposing the $m$ features (\eg~given $m=12$ and $N=3$, $SF_k$(A)~=~$T$($F_{4k}$(A), $F_{4k+1}$(A), $F_{4k+2}$(A), $F_{4k+3}$(A)), where $0{\leq}k{\leq}2$).

Using multiple SFs as the sketch of a data block enables SFSketch to tune the accuracy of reference search by changing the \rev{\emph{matching criteria} (i.e., criteria for judging the similarity of given two blocks)}.
Consider the example of \fig{\ref{fig:lsh}} where block A's data content is almost the same as block B, except for the red regions marked as \emph{Delta}.
Suppose that, due to the small differences between blocks A and B, every hash function $H_i$ other than $H_1$ has the same maximum value $Max(H_i(W_j))$ for blocks~A~and~B, \ie~every feature $F_i$ except $F_1$ is identical between blocks A and B.
In such a case, $SF_0$(A) does not match $SF_0$(B) ($\because F_1(A) \neq F_1(B)$), while all the other SFs are identical between blocks A and B.
The two blocks can be considered either similar or not depending on the matching criteria;
SFSketch may either decide that the two blocks are dissimilar because there exists a different SF or judge that block A resembles block B since their other two SFs (\ie~$SF_1$ and $SF_2$) match each other.
There are many possible matching criteria, but to maximize the data-reduction ratio, existing SFSketch-based techniques~\cite{zhang-fast-2019, shilane-fast-2012} consider that two blocks are similar if there exists at least one matching SF.

While existing SFSketch-based delta-compression techniques provide significant improvement in data reduction compared to a simple combination of deduplication and lossless compression~\cite{zhang-fast-2019, shilane-fast-2012}, we observe that SFSketch-based reference search often fails to identify a good reference block that can provide a high data-reduction ratio for an incoming block.
To show this, we compare a state-of-the-art SFSketch-based reference search technique~\cite{zhang-fast-2019} to \emph{brute-force} search that performs delta compression of an incoming block with \emph{every} stored block and selects the stored block that provides the highest data-reduction ratio as the incoming block's reference.\footnote{
While brute-force search guarantees the highest data-reduction ratio for a workload, it is infeasible to use due to its prohibitively high performance overhead. 
For example, in our evaluation environments (see \sect{\ref{subsec:methodology}} for more detail), brute-force search takes more than 300 hours for the \inst trace that writes a total of 8.83-GB data to the storage system.}  
For our evaluation, we use \fix{4,090,975} 4-KiB data blocks collected from six different workloads in real systems (see \sect{\ref{subsec:methodology}} for our evaluation methodology and workloads).

We use two major metrics to evaluate the accuracy of SFSketch compared to brute-force search: 1) \emph{false-negative rate (FNR)}, the probability of identifying no reference block for an incoming data block even though brute-force search can find one, and 2) \emph{false-positive rate (FPR)}, the probability of identifying a reference block \emph{different} from what brute-force search finds.  
For FN cases, SFSketch compresses the data block using the LZ4 algorithm~\cite{collet-lz4} because there is no reference block.  
For FP cases, SFSketch uses the Xdelta algorithm~\cite{macdonald-phdthesis-2000, macdonald-xdelta} to perform delta compression of the block with the reference block that it identifies.  
For both cases, we measure the average data-reduction ratio (DRR) and compare it with that of brute-force search. 
\tab{\ref{tab:lsh}} shows FNR, FPR, and DRR for FN/FP cases of the SFSketch-based reference search.
DRR is normalized to that of brute-force search.

\begin{table}[!h]
    \vspace{-.5em}
	\centering
	\caption{Accuracy of LSH-based reference search~\cite{zhang-fast-2019}.}
	\vspace{-1em}
	\resizebox{\columnwidth}{!}{%
	\begin{tabular}{c;c?p{.85cm}<{\centering}p{.85cm}<{\centering}p{.85cm}<{\centering}p{.85cm}<{\centering}p{.85cm}<{\centering}p{.85cm}<{\centering};p{.85cm}<{\centering}}
	\toprule
	\multicolumn{2}{c?}{\textbf{Workload}} & \textbf{\pc{}} & \textbf{\inst} & \textbf{\upd} & \textbf{\synth} & \textbf{\sensor} & \textbf{\web} & \textbf{Avg.}\\
	\hline
	\multicolumn{2}{c?}{\textbf{FNR}} & 35.3\% & 51.8\% & 56.3\% & 75.5\% & 48.1\% & 5.5\% & 35.7\%\\
	\multicolumn{2}{c?}{\textbf{FPR}} & 21.1\% & 15.8\% & 11.3\% & 14.1\% & 47.3\% & 60.6\% & 23.1\% \\
	\multirow{2}{*}{\textbf{DRR}} & \textbf{FN cases} & 0.474 & 0.488 & 0.578 & 0.639 & 0.567 & 0.539 & 0.562 \\
	 & \textbf{FP cases} & 0.621 & 0.608 & 0.644 & 0.683 & 0.798 & 0.674 & 0.669 \\ 
	\bottomrule
	\end{tabular}}
	\vspace{-.5em}
	\label{tab:lsh}%
\end{table}%

We make three observations from \tab{\ref{tab:lsh}}.  
First, the existing SFSketch-based technique suffers from high \om{FNR} (up to 75.5\% and 35.7\% on average), failing to find \emph{any} reference block for many incoming blocks that actually have one or more reference blocks.
Except for \web, SFSketch's \om{FNR is} higher than 35\% for every workload.
For FN cases, each data block is compressed by the LZ4 algorithm, and thus its DRR is considerably lower compared to when the block is delta-compressed with the reference block identified by brute-force search.
As shown in \tab{\ref{tab:lsh}}, the normalized DRR in FN cases is 0.562 on average, showing that SFSketch provides 43.8\% lower data reduction compared to the optimal for the FN cases (\ie~for 35.7\% of the entire data blocks on average).

Second, the SFSketch-based reference search frequently chooses a sub-optimal reference in some workloads, \eg~\sensor and \web, which have \om{a FPR} of 47.3\% and 60.6\%, respectively.
The sub-optimal selection of reference blocks results in lower data-reduction ratios over brute-force search.
As shown in the last row in \tab{\ref{tab:lsh}}, the normalized DRR in FP cases is 0.669 on average, which means that SFSketch provides 33.1\% lower data reduction compared to the optimal for the FP cases (\ie~for 23.1\% of the entire data blocks).

Third, FN cases are more common and have more negative impact on the DRR than FP cases.
On average, FN cases occur for 35.7\% of the incoming blocks, whereas FP cases occur for 23.1\%.
When a FN case happens, the data-reduction ratio using LZ4 is lower than when an FP case happens, which still uses delta compression albeit with a sub-optimal reference block; on average, the normalized DRR in FP cases (0.669) is 19\% higher than that in FN cases (0.562).

The limited accuracy of SFSketch mainly stems from its inherent property; SFSketch is highly optimized to identify \emph{only very similar} data.
Considering the SF-based sketching process explained in \fig{\ref{fig:lsh}}, it is highly unlikely that two blocks have at least one matching SF, unless they are very similar.  
This property enables SFSketch to provide a high data-reduction ratio even when it selects a sub-optimal reference block for an incoming block (\ie~for FP cases).
However, it also \om{causes} SFSketch \om{to} frequently fail to find a \textit{sufficiently good} reference block that is not very similar to the incoming block but is still beneficial for improving the data-reduction ratio.

It is challenging to optimize existing SF-based sketch algorithms to increase both FPR and FNR at the same time.
The accuracy of SFSketch highly depends on its settings such as the number of features ($m$), the number of super features ($N$), the sliding window size ($w$), and the matching criteria.
For example, under a matching criterion where two blocks are considered similar if they have at least one common SF, increasing the number of SFs \sj{(\ie~$N$)} for each data block would reduce overall FNR.
However, at the same time, it might increase FPR and reduce data-reduction ratios in FP cases because more dissimilar blocks could be chosen as reference blocks. 
Moreover, as shown in \tab{\ref{tab:lsh}}, the FNR/FPR trend of SFSketch-based search greatly varies across workloads, which make\om{s} it even more difficult to find the best configuration on average as well as on a per-workload basis.
Instead, we investigate applicability of deep-learning algorithms for data sketching in delta compression, which can reduce the human effort for developing a new sketching scheme or fine-tuning existing techniques for different workloads.

\vspace{-1em}
\subsection{Learning-to-Hash Method \label{subsec:motiv_l2h}}
\vspace{-.5em}
The learning-to-hash method~\cite{kulis-neurips-2009, wang-ieeetpami-2017} is a promising machine learning (ML)-based approach for the nearest-neighbor search problem.
It trains a neural network (NN) to generate a hash value for a given input data block such that any two similar data blocks have similar hash values.
Many prior works~\cite{zhu-AAAI-2016, cao-AAAI-2017, leng-icml-2015, lin-cvpr-2015, erin-cvpr-2015, cao-cvpr-2018, liang-atc-2019} demonstrate the high effectiveness of the learning-to-hash method at nearest-neighbor search in various applications, such as image recognition and classification.

\begin{figure}[!b]
    \vspace{-1em}
    \centering
    \includegraphics[width=\linewidth]{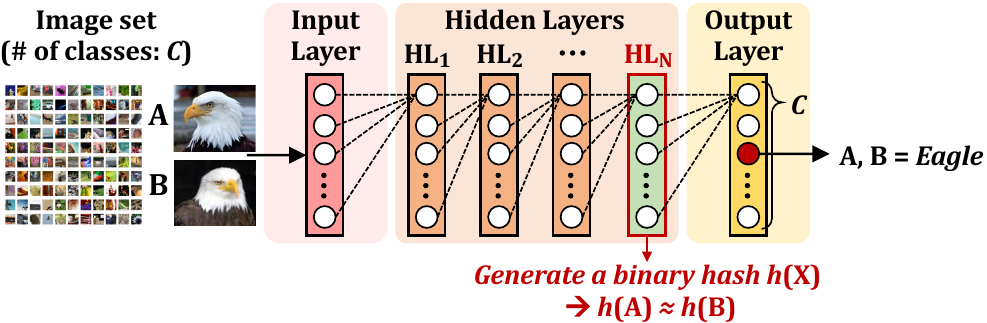}
   \vspace{-2em} 
    \caption{Learning-to-hash for image retrieval.}
    \label{fig:l2h}
    \vspace{-.3em} 
\end{figure}

\fig{\ref{fig:l2h}} depicts how a representative learning-to-hash scheme~\cite{lin-cvpr-2015} generates a binary hash value of an image for
content-based image retrieval.  
It extracts the hash value of an input image from the last hidden layer (\eg~$\text{HL}_\text{N}$ in \fig{\ref{fig:l2h}}) of \om{a} NN that is trained to classify the input image to one of $C$ possible classes.
During inference, the activations in the last hidden layer of two similar images are likely to be largely the same if the two images belong to the same class.
Therefore, their hash values should also be similar because they are directly extracted from the last layer by translating the output of each activation into a binary (`1' or `0').

The learning-to-hash method has potential to be used for reference block search in post-deduplication delta compression, another nearest-neighbor search problem.
In particular, rapid advances \om{in} machine learning have enabled learning-based algorithms to outperform a human or human-made heuristics in various problems, such as facial recognition~\cite{phillips-procnas-2018}, speech recognition~\cite{xiong-archieve-2016, xiong-ieeetaslp-2017}, image classification~\cite{lin-cvpr-2015}, and system optimizations (\eg~branch prediction~\cite{jimenez-hpca-2001, jimenez-micro-2003}\rev{,} memory scheduling~\cite{ipek-isca-2008}\om{, \rev{and} prefetching~\cite{bera-micro-2021}}).
These successful examples motivate us to develop a learning-based sketching scheme that could be more effective than existing LSH-based sketching schemes relying on human-designed heuristics and metrics.
\vspace{-1em}
\section{\prop\label{sec:deepsketch}}
\vspace{-.5em}
The key idea of \prop is to use the learning-to-hash method to generate similar data signatures (\ie~sketches) for any two data blocks that would provide a high data-reduction ratio when delta-compressed relative to each other.
The main difference of \prop over the existing post-deduplication delta-compression approach~\cite{shilane-fast-2012, zhang-fast-2019} (described in \fig{\ref{fig:baseline}}) is that \prop generates a data block's sketch by using a deep neural network (DNN) model, instead of using an LSH function (\eg~$sk$(X) in \fig{\ref{fig:baseline}}).
For each incoming data block, \prop performs DNN inference with the block as input and uses the resulting activation values \om{in} the DNN's last hidden layer as the block's sketch.

We envision that \prop's DNN is \emph{pre-trained} offline in other machines with more powerful computation resources before building or updating a storage machine.
For example, to adopt \prop in a new storage server of a data center, one can train \prop's DNN using randomly-selected data blocks from existing storage servers that contain specific types of data (\eg~databases, images, web caching, etc.) expected to be stored in the new storage server.
Similarly, to further enhance the accuracy of \prop, one can retrain \prop's DNN and use the enhanced DNN to build new storage servers or reorganize existing ones.

While the high-level idea may sound simple, applying the learning-to-hash method for reference search in post-deduplication delta compression is not straightforward.
This is because \prop needs to deal with \emph{general binary data}, which introduces the following two technical challenges:
\linebreak
\head{Challenge 1. Lack of Semantic Information} The target data set of \prop can contain \emph{any} data from various applications, such as text, images, binary executable files, and so on.  
Compared to existing learning-to-hash approaches focusing on pre-categorized data (\eg~Imagenet~\cite{deng-cvpr-2009}, CIFAR~\cite{krizhevsky-2009}, and MNIST~\cite{lecun-1998}), \prop needs to process a much wider range of data without any well-defined semantic information about the delta-compressibility of data blocks.

\head{Challenge 2. High Dimensional Space} 
The lack of semantic information in \prop's target data sets leads us to perform \emph{unsupervised learning} that is used for drawing inferences from a data set without labeled information.  
The most common unsupervised learning approach is to use a clustering algorithm that groups the target data set according to a certain measure, \eg~similarity of bit patterns in our case.
However, possible bit patterns of a data block for \prop have extremely high dimensional space (\eg~2$^{4,096\times8}$ assuming a 4-KiB data block), which makes it difficult to 1) set a proper number of final clusters and 2) collect a data set large enough to cover all possible data patterns for a clustering algorithm.

To address the above challenges, we develop a new clustering algorithm, called \emph{dynamic k-means clustering (\cluster)}, which groups data blocks that would provide high delta-compression ratios when delta-compressed relative to each other (\sect{\ref{subsec:clustering}}). 
To cope with potential groups of data blocks that are missing in the collected data sets, after clustering, we generate new data blocks by randomly and slightly modifying existing blocks.
We then generalize the understanding of the similarity relationship between data blocks using the learning-to-hash method, so that \prop can generate a learning-based data sketch for any given block (\sect{\ref{subsec:training}}).
With the sketch values computed by the learning-to-hash model, \prop identifies the most similar reference block to each incoming block based on an approximate nearest-neighbor search technique (\sect{\ref{subsec:select}}).
We also perform hyper-parameter exploration for our DNN model to find the appropriate sketch size (\sect{\ref{sec:exploration}}).

\vspace{-1em}
\subsection{Dynamic K-Means Clustering\label{subsec:clustering}}
\vspace{-.5em}
\cluster is based on the existing k-means clustering algorithm~\cite{lloyd-ieeetit-1982} that partitions a data set into a given number (\ie~$k$) of clusters such that each data element belongs to the cluster with the nearest mean value.
Unfortunately, in our case, the value of $k$ is initially unknown.
Figuring out the most suitable value for $k$ by exploring a given data set is 
time-consuming, considering the extremely high dimensionality of the data set that \prop deals with.

The hierarchical clustering algorithm~\cite{johnson-psychometrika-1967} is known to be suitable for such data sets, but it introduces prohibitive computation and memory overheads for a large-size data set.
To be specific, the computation and space complexities of hierarchical clustering are $O(N^3)$ and $O(N^2)$, respectively, where $N$ is the number of data blocks to cluster.
This means that, for example, hierarchical clustering of a 4-GB data set requires TB-scale memory space assuming a data block size of 4 KiB.

There exist a number of \emph{adaptive} clustering algorithms (\eg~\cite{shafeeq-icicn-2012, bhatia-flairs-2004, zheng-eurasipjivp-2018, nayini-icecds-2017, hamerly-neurips-2003}) that aim to cluster a data set with limited knowledge of the number of final clusters.
Unfortunately, using them for DNN training in \prop is not straightforward either, because their efficiency also highly depends on the initial parameters that are set either randomly or manually, such as the initial number of clusters~\cite{shafeeq-icicn-2012, zheng-eurasipjivp-2018, nayini-icecds-2017, hamerly-neurips-2003} or the distance threshold to determine the similarity of given two objects~\cite{bhatia-flairs-2004}.
Since the target data set of \prop has an extremely high dimensional space while there is no available hint for good initial parameters, using existing techniques could either require significant effort to find appropriate initial parameters or lead to limited accuracy and/or prohibitive performance overhead due to the use of inappropriate initial parameters.   

\begin{figure}[!h]
    \centering
    \vspace{-1em}
    \includegraphics[width=\linewidth]{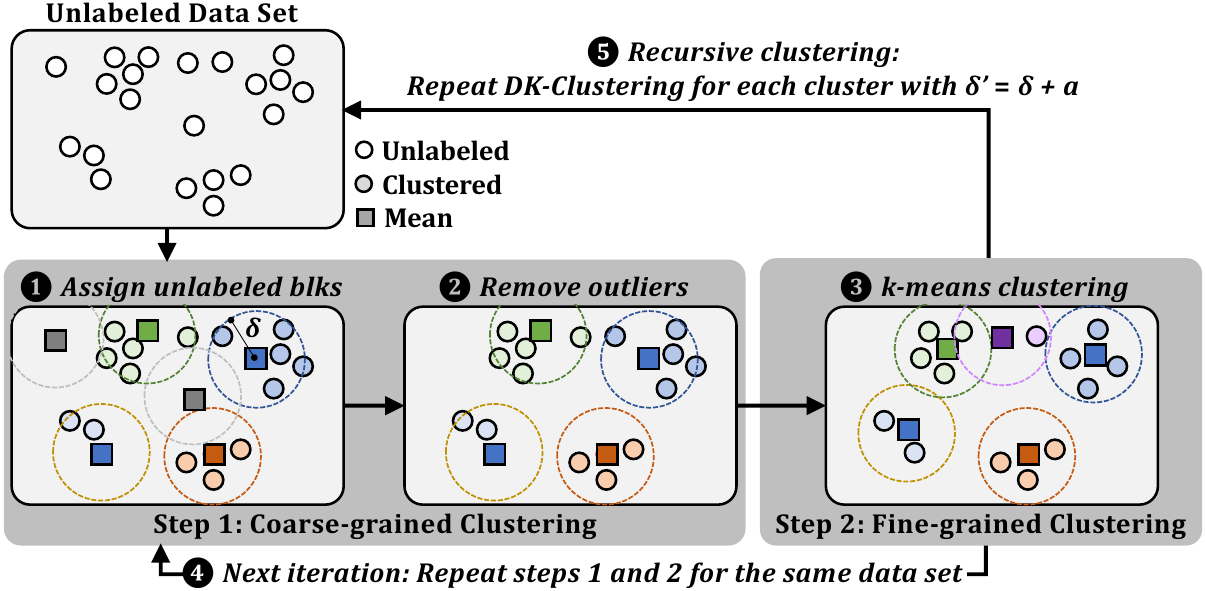}
    \vspace{-1.5em}
    \caption{Overall procedure of dynamic k-means clustering.}
    \vspace{-1em}
    \label{fig:clustering}
\end{figure}

To overcome the above challenges, we develop \cluster{} by extending the existing k-means clustering algorithm with specialized initialization steps to dynamically refine the value of $k$ while clustering data without any hints for initial parameters.
\fig{\ref{fig:clustering}} describes the overall process of \cluster composed of two
steps that are performed iteratively: \textbf{Step 1.} coarse-grained clustering and \textbf{Step 2.} fine-grained clustering.  
Coarse-grained clustering first creates rough clusters within an unlabeled data set, and then fine-grained clustering adjusts the assignment of data blocks by running a modified k-means clustering algorithm. 
Fine-grained clustering returns a data block to be unlabeled if the block is an \emph{outlier} in the cluster, so that coarse-grained clustering can re-categorize the block at the next iteration.
After Steps 1 and 2 converge, \cluster repeats the above steps for \emph{each} cluster in a recursive manner, which enables us to form fine-grained clusters that only contain data blocks sufficiently similar to each other.

\head{Step 1: Coarse-Grained Clustering}
Coarse-grained clustering takes a set of unlabeled blocks and clusters as the input, and aims to categorize all the unlabeled blocks.
Initially, there exist only unlabeled blocks but no cluster, so \cluster{} creates a new cluster and assigns the first block as the representative block (\ie~\emph{mean}) of the cluster. 
After that, for each unlabeled block, \cluster{} measures the data-reduction ratio when the block is delta-compressed with the mean of each cluster through the target delta-compression algorithm (\eg~Xdelta~\cite{macdonald-phdthesis-2000, macdonald-xdelta}).  
\cluster{} selects the cluster whose mean provides the highest data-reduction ratio for the unlabeled data block.  
If the data-reduction ratio is higher than a threshold $\delta$, \cluster{} adds the unlabeled block to the selected cluster.
Otherwise, it creates a new cluster, and the unlabeled block becomes the new cluster's mean (\bcirc{1} in Figure~\ref{fig:clustering}). 
After categorizing all unlabeled blocks, coarse-grained clustering removes clusters that contain only a single data block from the data set as there are likely no other blocks sufficiently similar to that block (\bcirc{2}). 

\head{Step 2: Fine-Grained Clustering} 
Since coarse-grained clustering roughly assigns unlabeled blocks to clusters, it cannot guarantee that all the data blocks belonging to the same cluster are sufficiently similar to each other.  
To address this, \cluster performs fine-grained clustering for the resulting clusters from coarse-grained clustering.
Fine-grained clustering performs a variant of k-means clustering, adjusting the mean of each cluster and re-assigning each data block to the cluster containing the nearest mean (\bcirc{3}).
Fine-grained clustering operates differently from the typical k-means clustering in \js{three} aspects.
First, instead of Euclidean distance~\cite{danielsson-cgip-1980}, it uses the delta-compression ratio of two data blocks as the distance function.
Second, it derives a cluster's mean by \emph{selecting} the block that provides the highest average data-reduction ratio when delta-compressed relative to each of the other blocks in the cluster.
Third, if there is a data block whose delta-compression ratio when delta-compressed relative to the cluster's mean is lower than the threshold $\delta$, \cluster{} excludes the block from the cluster and considers it as an unlabeled block.
After finishing fine-grained clustering on all the clusters, \cluster repeats Steps 1 and 2 over all the clusters until no unlabeled data blocks exist (\bcirc{4}).

\head{Step 3: Recursive Clustering} 
Fine-grained clustering guarantees that \emph{every} resulting data block belongs to an appropriate cluster where the data block provides a data-reduction ratio higher than the given threshold $\delta$ when delta-compressed relative to the cluster's mean.
Even though a sufficiently high value for $\delta$ would allow \cluster{} to group only similar data blocks into the same cluster, other values for $\delta$ can provide better clustering results. 
In order to automatically find the best $\delta$ for a data set, once \cluster{} reaches the convergence with a given threshold $\delta$, it performs Steps 1 and 2 for \emph{each} cluster using a new threshold $\delta' = \delta + \alpha$ in a recursive manner (\bcirc{5}).
Data blocks assigned to each cluster are considered unlabelled again for the next recursion with the new threshold $\delta'$. 
The recursion terminates when splitting a cluster shows no more benefit in improving the data-reduction ratio.
More specifically, \cluster stops the recursion for a cluster if the average data-reduction ratio of data blocks in the cluster is similar or lower than the average ratio of sub-clusters spawned from the cluster.

\head{\cluster{} Complexity}
The space complexity of DK-Clustering is $O(N)$ since it only requires storing per-block information about which cluster the block belongs to.  
The computation complexity of \cluster{} is $O(N \times K_F) + O(N^2/K_C) < O(N^2)$, where $K_C$ and $K_F$ are the number of total clusters after coarse-grained and fine-grained clustering steps, respectively.
Although the number of iterations for \cluster{} can vary depending on workload, \cluster{} finishes within up to eight iterations for our training data sets.
Note that, even for an extreme case where \cluster{} requires a large number of iterations, we can easily limit the maximum number of iterations at minimal degradation in clustering quality. 
For example, one can set a threshold distance to finish \cluster{} once it groups all data blocks such that any data block’s distance from the corresponding cluster’s mean is lower than the threshold distance.

\vspace{-1em}
\subsection{Neural-Network Training\label{subsec:training}}
\begin{figure}[!t]
    \centering
    \includegraphics[width=1.00\linewidth]{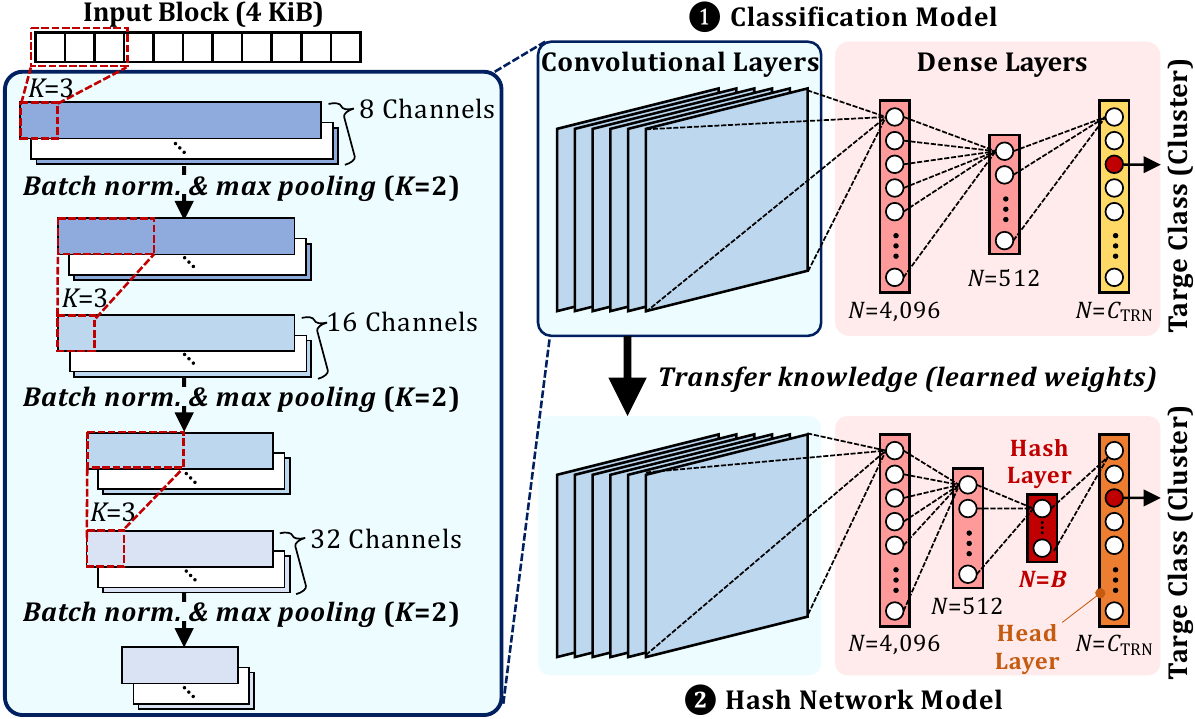}
    \vspace{-1.5em}
    \caption{NN models of \prop.}
    \label{fig:model}
    \vspace{-1em}
\end{figure}

\vspace{-.5em}
\fig{\ref{fig:model}} shows our method to train a DNN model for \prop to generate a data block's sketch, which consists of two steps.
In the first step, we train a \emph{classification} model (\bcirc{1}) 
using the $C_\text{TRN}$ clusters formed by \cluster as different target classes.
The first part consists of three standard 1D convolutional layers applying the max pooling and batch normalization techniques, which capture spatial locality of neighbor bytes within the data block.
The network is then connected to dense layers to learn the relationship between the extracted spatial features and the target class.\footnote{
We explore multiple NN structures and choose the one that provides the best classification accuracy and data-reduction ratio (shown in \fig{\ref{fig:model}}). 
For example, when using a much simpler multi-layer perceptron (MLP) networks~\cite{gardner-ae-1998}, \prop hardly provides data-reduction benefits (less than 1\%) over existing SF-based techniques. 
Adding the number of dense layers in the classification model in \fig{\ref{fig:model}} does not improve classification quality, either. 
We discuss detailed results for hyper-parameter search in~\sect{\ref{sec:exploration}}.}

After training the classification model, in the second step, we transfer the learned knowledge of the classification model to a \emph{hash} network model (\bcirc{2}).
We employ a state-of-the-art learning-to-hash technique called GreedyHash~\cite{su2018greedy}. 
We first initialize the hash network with the weights of the classification model.
Instead of using the last layer of the classification model, we train the hash network with two different layers, a hash layer and a head layer, each of which learns the binary hash and class likelihood, respectively.

A key challenge in NN training for \prop is that data blocks are not uniformly distributed over $C_\text{TRN}$ clusters.
In our data set, the largest 10\% clusters contain 47.93\% of the total data blocks.
It would render training of the NN to be significantly biased towards specific bit patterns.
To address this, we resize each of $C_\text{TRN}$ clusters to have the same number of $N_\text{BLK}$ blocks by 1) randomly selecting $N_\text{BLK}$ blocks within a cluster containing more blocks than other clusters and 2) adding data blocks randomly and slightly modified from ones in a cluster containing fewer blocks.

Once training the hash network, the hash layer yields the $B$-bit representation for an input block, \ie~the input block's sketch, allowing any two similar data blocks to have similar sketches with low Hamming distance.
Note that, even if two data blocks do not belong to any of $C_\text{TRN}$ clusters, we can infer their binary hash values based on the likelihood that each block belongs to the clusters, which dramatically improves the adaptability of our NN model over various data sets.

\vspace{-1em}
\subsection{Reference Selection\label{subsec:select}}
\vspace{-.5em}
\prop identifies whether or not any two given data blocks are similar by comparing the two blocks' sketches generated from the hash network model.
A key challenge here is that the traditional exact-matching-based search method (which uses a hash table for the SK store) is not effective for the learning-to-hash model.
For example, the hash network model may generate similar but few-bit different sketches for some blocks beneficial to be delta-compressed, which causes an exact-matching-based search method to misjudge those blocks to be dissimilar.

To address this issue, we use the approximate nearest-neighbor search (ANN) technique.
Unlike the standard exact nearest-neighbor search, ANN techniques provide a scalable and performance-efficient way to find the most similar values by relaxing search conditions.
In particular, we use the NGT library~\cite{ngt} that supports searching with high-dimensional binary data using neighborhood graphs and tree indexing.

\begin{figure}[!h]
    \centering
    \vspace{-.5em}
    \includegraphics[width=\linewidth]{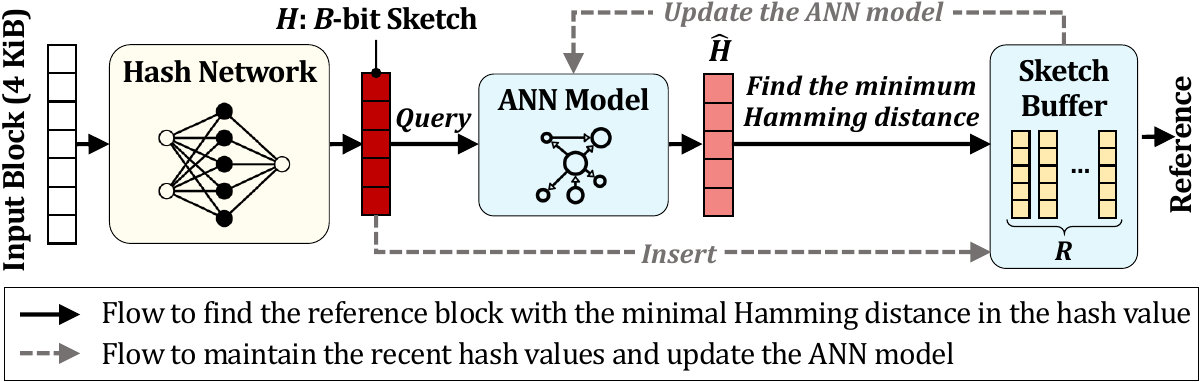}
    \vspace{-1.5em}
    \caption{Overview of the reference selection procedure.}
    \label{fig:refflow}
    \vspace{-.5em}    
\end{figure}

\fig{\ref{fig:refflow}} illustrates the reference selection procedure of \prop.
For each incoming block, \prop first computes its sketch, $\mathbf{H}$, using the hash network model.
It then searches for the similar block from two SK stores.
The first SK store utilizes the ANN technique, and \prop queries it with $\mathbf{H}$ to get the data block with the most similar sketch, $\widehat{\mathbf{H}}$, in the ANN model.
The other SK store buffers the sketches of $R$ most-recently-written blocks.
Let $\Delta(\mathbf{H}, \widehat{\mathbf{H}})$ be the Hamming distance between the two hash values.
For each recent block in the buffer store, \prop checks if there is a block with a Hamming distance smaller than  $\Delta(\mathbf{H}, \widehat{\mathbf{H}})$.
If there exists, \prop chooses the block from the buffer store as the reference for the incoming block. 
Otherwise, it uses the block from the ANN-based SK store (\ie~the block whose sketch is $\widehat{\mathbf{H}}$) as the reference.

The underlying reason for using the two SK stores is that, under the current implementation using the NGT library, updating the ANN model takes a non-negligible amount of time.
To avoid frequent updates of the data structure that would hurt the performance of \prop, we design \prop to update the ANN model in a batch by buffering the sketches of recently-written data blocks.
When the number of sketches in the buffer exceeds a threshold \tblk (\eg~128 in our default settings), \prop flushes the buffered sketches to the ANN-based SK store.
Note that it is important to check the sketch buffer in order to maximize the data-reduction ratio of \prop. 
In our evaluation, 13.8\% of the reference blocks are found in the sketch buffer on average (up to 33.8\%).

\vspace{-1em}
\subsection{Hyper-Parameter Exploration\label{sec:exploration}}
\vspace{-.5em}
This section presents our hyper-parameter exploration for \prop to achieve high accuracy in reference search with a convolutional hash network.

\head{Classification Model}
As discussed in~\sect{\ref{subsec:training}}, the DNN training procedure of \prop has two steps to train the classification model and the hash network model, respectively.
To generate accurate sketches of data blocks, the classification model should predict the correct target classes (\ie~the clusters formed by \cluster).
We identify the best hyper-parameters for the proposed classification model using the standard machine learning practice of the grid search along with nested cross-validation.
We choose the number of the convolutional and dense layers from the grid $\langle1,2,3\rangle$, the number of the convolution channels size from $\langle8,16,32,64\rangle$, the number of neurons for each dense layer from $\langle512, 1,024, 2,048, 4,096\rangle$, the dropout rate for the dense layers from $\langle0.0, 0.1, 0.2, 0.5\rangle$, and the learning rate from $\langle0.01, 0.02, 0.005, 0.1, 0.5\rangle$.
We utilize ReLU for the activation function for each layer and train the model with the Adam optimizer~\cite{kingma-iclr-2015}.
We use 10\% of samples in our data sets for training and the remaining 90\% for testing.
Finally, we select the proposed classification model structure that shows the best testing accuracy in the cross-validation.

\begin{figure}[!h]
    \centering
    \vspace{-.5em}
    \includegraphics[width=0.8\linewidth]{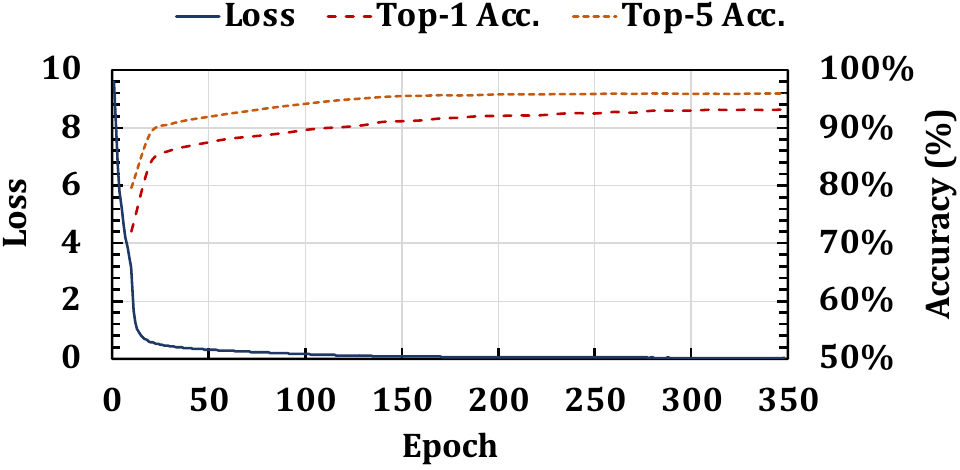}
    \vspace{-1em}
    \caption{Loss and accuracy of classification model.}
    \vspace{-.5em}
    \label{fig:hploss}
\end{figure}

\fig{\ref{fig:hploss}} shows the loss and testing accuracy changes over training epochs for the classification model.
The proposed classification model accurately predicts the target cluster identified in \cluster even though the data sets used in our evaluation has a relatively large number of the clusters, $C_\text{TRN}=34,025$.
After training with 350 epochs, the model training procedure sufficiently converges, achieving $93.42\%$ for Top-1 and $96.02\%$ for Top-5 accuracy.
It implies that the deep learning method itself can accurately identify similar blocks for an incoming block without any other information.

\head{Hash Network Model}
Next, we train the hash network model while changing the sketch size $B$. 
With the smaller~$B$, similar data blocks would have a higher chance to have the same hash value, but it also increases the false-positive rate, \ie~dissimilar blocks belonging to different clusters would have the same hash value.
One may set the number of bits with a sufficiently large number, but doing so increases the memory overhead for the SK store and the computation time for ANN search and update processes.

\begin{figure}[!h]
    \centering
    \includegraphics[width=\linewidth]{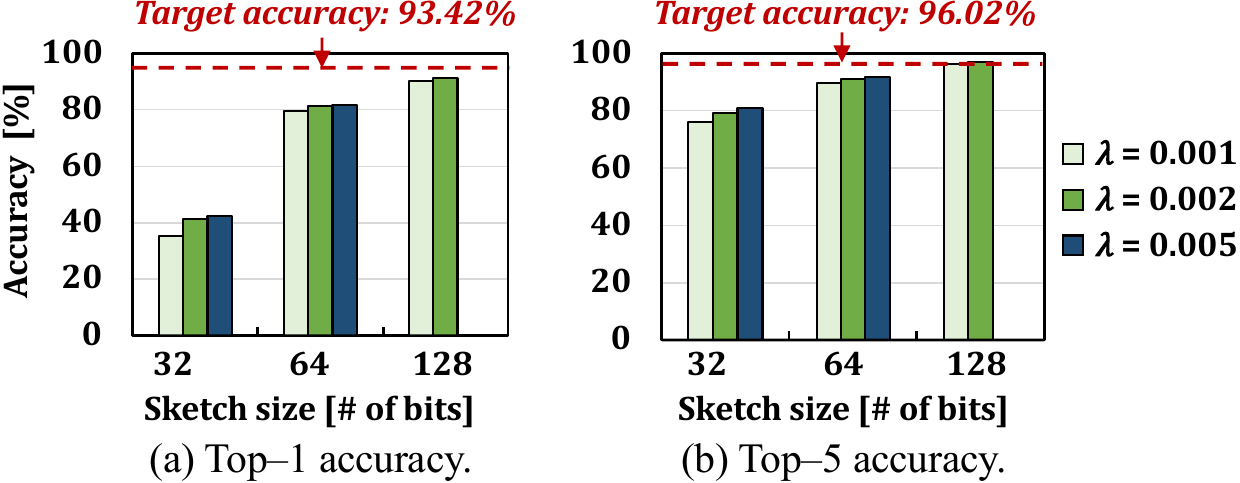}
    \vspace{-2em}
    \caption{Accuracy of hash network model.}
    \vspace{-1em}
    \label{fig:acc}
\end{figure}

To determine the best sketch size, we verify when the hash network model could achieve the classification model's original accuracy.
Recall that the hash network model learns both the hash coding and classification at the same time.
Thus, we can verify whether it correctly classifies the target class by checking the last head layer's activations.
\fig{\ref{fig:acc}} summarizes our evaluation results.
We evaluate three candidate values for $B$, 32, 64, and 128, over different learning rates $\lambda$.
Note that the model does not converge when $B=128$ and $\lambda=0.005$, so we omit the results.
The results show that the hash network model does not recover the accuracy of the classification model with the small hash bits, $32$ and $64$, since the representation capability of the hash coding is insufficient.
When $B=128$, we observe that the hash network model also predicts the target clusters with a high accuracy, \eg~it achieves the Top-5 accuracy of 96.92\% with $\lambda=0.002$, exceeding the original target accuracy of the classification model.
Thus, we decide to use $B=128$ for our implementation of \prop.

\vspace{-1em}
\section{Evaluation\label{sec:evaluation}}
\vspace{-.5em}
In this section, we evaluate the data-reduction benefits and performance/memory overheads compared to the state-of-the-art super feature (SF)-based sketching technique~\cite{zhang-fast-2019}.

\vspace{-1em}
\subsection{Methodology\label{subsec:methodology}}
\vspace{-.5em}
\head{Evaluation Platform}
We develop a post-deduplication delta-compression platform that is used as a general workbench to implement and evaluate various reference search techniques.\footnote{We open source our platform along with the data sets used in our evaluation~\cite{deepsketch-git}.}
Our platform runs on a server machine that employs Intel's Xeon 4110 CPU with 8 cores running at 2.1~GHz, 128-GB DDR4 DRAM, and 8 Samsung 860PRO 1-TB SSDs, 
while using GeForce RTX 2080 for DNN inference in \prop.

Our platform operates as described in \fig{\ref{fig:baseline}}; for every host write, it performs deduplication, delta compression, and lossless compression in order.  
It maintains three main data structures: 1) a fingerprint store for deduplication, 2) a sketch store for delta compression, and 3) a reference table for serving future read requests.
The data block size is 4 KiB, which is identical to the default block size of widely-used file systems~\cite{mathur-linux-2007, custer-microsoft-1994}.
We use the MD5 cryptographic hash algorithm~\cite{rivest-1992} to generate a 128-bit fingerprint of an incoming data block and the Xdelta delta-compression algorithm~\cite{macdonald-phdthesis-2000, macdonald-xdelta} to compress a non-deduplicated data block with its reference block.  
If there are multiple reference blocks similar to an incoming data block, our platform uses the first-found one as a reference by default.
When the platform cannot find a reference block, it compresses the incoming block using the LZ4 algorithm~\cite{collet-lz4}.
We set the threshold for the number of buffered sketches to invoke ANN updates to 128, which we empirically determine to minimize the performance overhead of exhaustive search and prevent too frequent ANN updates.

\head{Baseline Technique}
We compare \prop against \js{\emph{Finesse}~\cite{zhang-fast-2019},} the state-of-the-art SF-based technique \js{that provides much higher throughput while retaining almost the same data-reduction ratio compared to the representative post-deduplication delta-compression technique~\cite{shilane-fast-2012}}.
We configure Finesse using the default settings presented in~\cite{zhang-fast-2019}, which are already optimized and have been shown to provide the best data-reduction efficiency with low overhead for a wide range of workloads.
Finesse generates three 192-bit SFs, each of which can be obtained by transposing four features from different hash functions (\ie~twelve (= 3 $\times$ 4) Rabin fingerprint functions~\cite{rabin-techreport-1981} with a window size of 48 bytes are used in total). 
It considers that two data blocks are similar if they have one or more matching SFs, and selects the data block that has the largest number of matching SFs with the incoming block as the reference block for delta compression.

\head{Workloads}
We use \fix{eleven} block I/O traces that we collect by running different applications on real systems and capturing write requests (including the requested data) to the storage devices. There is no backup process during trace collection.
\tab{\ref{tab:workloads}} summarizes the characteristics of the traces in terms of the size, deduplication ratio (\ie~\emph{Orignial Data-Set Size / Data-Set Size after Deduplication}), and average compression ratio (\ie~\emph{Original Data-Set Size / Compressed Data-Set Size}).
We collect the I/O traces including contents written in the storage system from real desktop machines and servers while running different applications.

\begin{table}[!h]
	\centering
	\vspace{-.5em}
	\caption{Summary of the evaluated workloads.}
	\vspace{-1em}
	\resizebox{1\columnwidth}{!}{%
	\begin{tabular}{ccccc}
		\toprule
		\multirow{2}{*}{\textbf{Workload}} & \multirow{2}{*}{\textbf{Description}} & \multirow{2}{*}{\textbf{Size}} & \textbf{Dedup.} & \textbf{Comp.} \\
		& & & \textbf{ratio} & \textbf{ratio} \\
		\midrule
		\textbf{\textsf{PC}} & General Ubuntu PC usage & 1.57 GB & 1.381 & 2.209 \\
		\hline
		\textbf{\textsf{Install}} & Installing \& executing programs & 8.83 GB & 1.309 & 2.45 \\
		\hline
		\textbf{\textsf{Update}} & Updating \& downloading SW packages & 3.73 GB & 1.249 & 2.116 \\
		\hline
		\textbf{\textsf{Synth}} & Synthesizing hardware modules & 653 MB & 1.898 & 2.083 \\
		\hline
		\textbf{\textsf{Sensor}} & Sensor data in semiconductor fabrication &  91.2 MB & 1.269 & 12.38 \\
		\hline
		\textbf{\textsf{Web}} & Web page caching & 959 MB & 1.9 & 6.84\\
		\hline
		\textbf{\textsf{SOF0}} &  & 8.98 GB & 1.007 & 2.088\\
		\hhline{-~---}
		\textbf{\textsf{SOF1}} & Storing Stack Overflow database~\cite{sof} & 13.6 GB & 1.01 & 1.997 \\
		\hhline{-~---}
		\textbf{\textsf{SOF2}} & as of 2010 (\sof{0}) and 2013 (\sof{1--4}) & 13.6 GB & 1.01 & 1.996 \\
		\hhline{-~---}
		\textbf{\textsf{SOF3}} &  & 13.6 GB & 1.01 & 1.997 \\
		\hhline{-~---}
		\textbf{\textsf{SOF4}} &  & 13.6 GB & 1.01 & 1.996 \\
		\bottomrule
	\end{tabular}}
	\label{tab:workloads}%
	\vspace{-.5em}
\end{table}%

For \prop, we use different sets of data for training and testing.
In order to evaluate the \emph{adaptability} of \prop (\ie~how well \prop operates under a workload totally different from ones used in its DNN training), we do \emph{not} use the five traces collected from Stack Overflow database~\cite{sof} (\sof{0--4}) for training the DNN of \prop.
By default, we train \prop's DNN model using a single data set that contains 10\% of all the remaining six traces and evaluate \prop with the remaining 90\% of the six traces and entire SOF traces.

\vspace{-1em}
\subsection{Overall Data Reduction\label{subsec:overall}} 
\begin{figure}[!h]
    \centering
    \vspace{-.5em}
    \includegraphics[width=\linewidth]{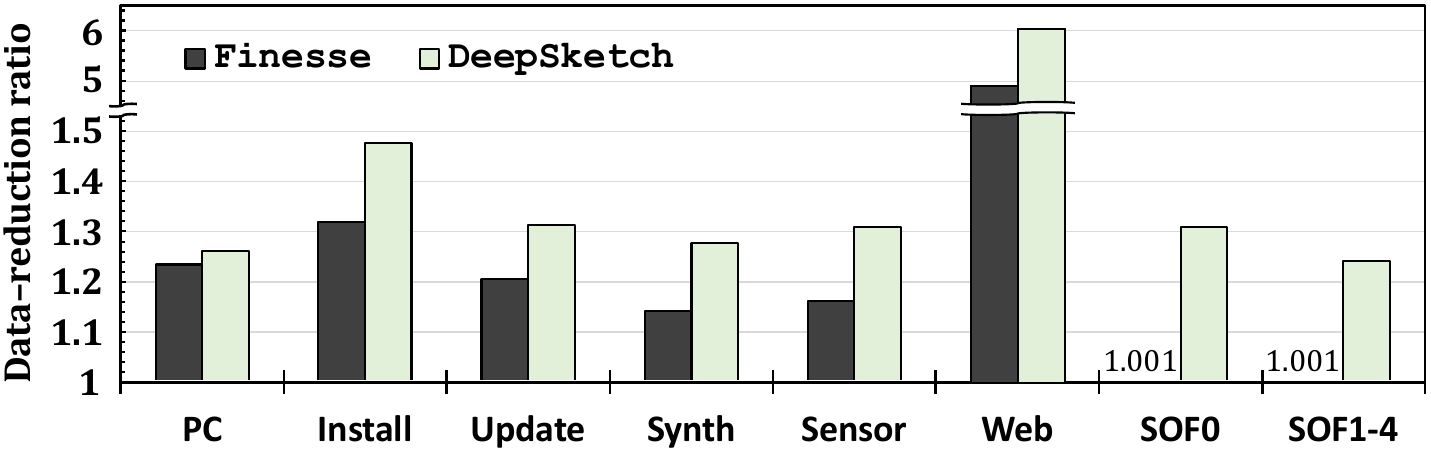}
    \vspace{-2em}
    \caption{Comparison of overall data-reduction ratio.}
    \label{fig:overall}
     \vspace{-1em}
\end{figure}

\vspace{-.5em}
\fig{\ref{fig:overall}} shows the data-reduction ratio after post-dedup\-lication delta compression with the two reference search techniques, Finesse and \prop, under the \fix{eleven} workloads.
We only present \sof{1} as a representative result of \sof{1-4} as they show little variations lower than 0.01\%.
All values are normalized to the data-reduction ratios of a baseline system that performs only deduplication and lossless compression in order, which we call \emph{no delta compression (noDC)}.  

We make \js{two} main observations from \fig{\ref{fig:overall}}.  
\js{First,} \prop significantly outperforms Finesse in most workloads.
Except for \pc{} in which \prop provides the similar data-reduction ratio with Finesse, \prop exhibits up to 33\% (on average 21\%) higher data-reduction ratios than Finesse. 
In particular, \prop greatly improves the data-reduction ratio by \emph{at least} 24\% over Finesse under \sof{} workloads.
This suggests that 1) \prop can improve the data-reduction efficiency for workloads that the state-of-the-art SF-based search technique cannot effectively cope with, and 2) \prop has high adaptability (\ie~it can work efficiently for data sets that are not used for the DNN training).
Second, \prop provides higher data-reduction ratios even for highly compressible workloads.
Under \web, Finesse significantly reduces the write traffic by about 80\% over noDC, but \prop increases the data-reduction ratio even further by 33\% compared to Finesse.  
From our observations, we conclude that \prop is an effective solution to maximize the data-reduction ratio for various workloads.

\vspace{-1em}
\subsection{Reference Search Pattern Analysis\label{subsec:pattern}}
\vspace{-.5em}
To better understand how \prop can outperform the state-of-the-art technique, we analyze the reference-search efficiency of \prop and Finesse.
Given a data block $B$, we measure $S_{FS}(B)$ and $S_{DS}(B)$, the number of \emph{saved bytes} by Finesse and \prop, respectively.
$S_{FS}(B)$ (or $S_{DS}(B)$) is obtained by subtracting the size of $B$ when delta-compressed with the reference block found by Finesse (or \prop) from the original size of $B$ (\ie~4 KiB).
The larger the $S_{FS}(B)$ (or $S_{DS}(B)$) value, the higher the reference-search efficiency. 
If a reference search technique fails to find a reference block for $B$, we compress it using the LZ4 algorithm and then use the compressed size to calculate data saving.

\begin{figure}[!t]
    \centering
    \includegraphics[width=\linewidth]{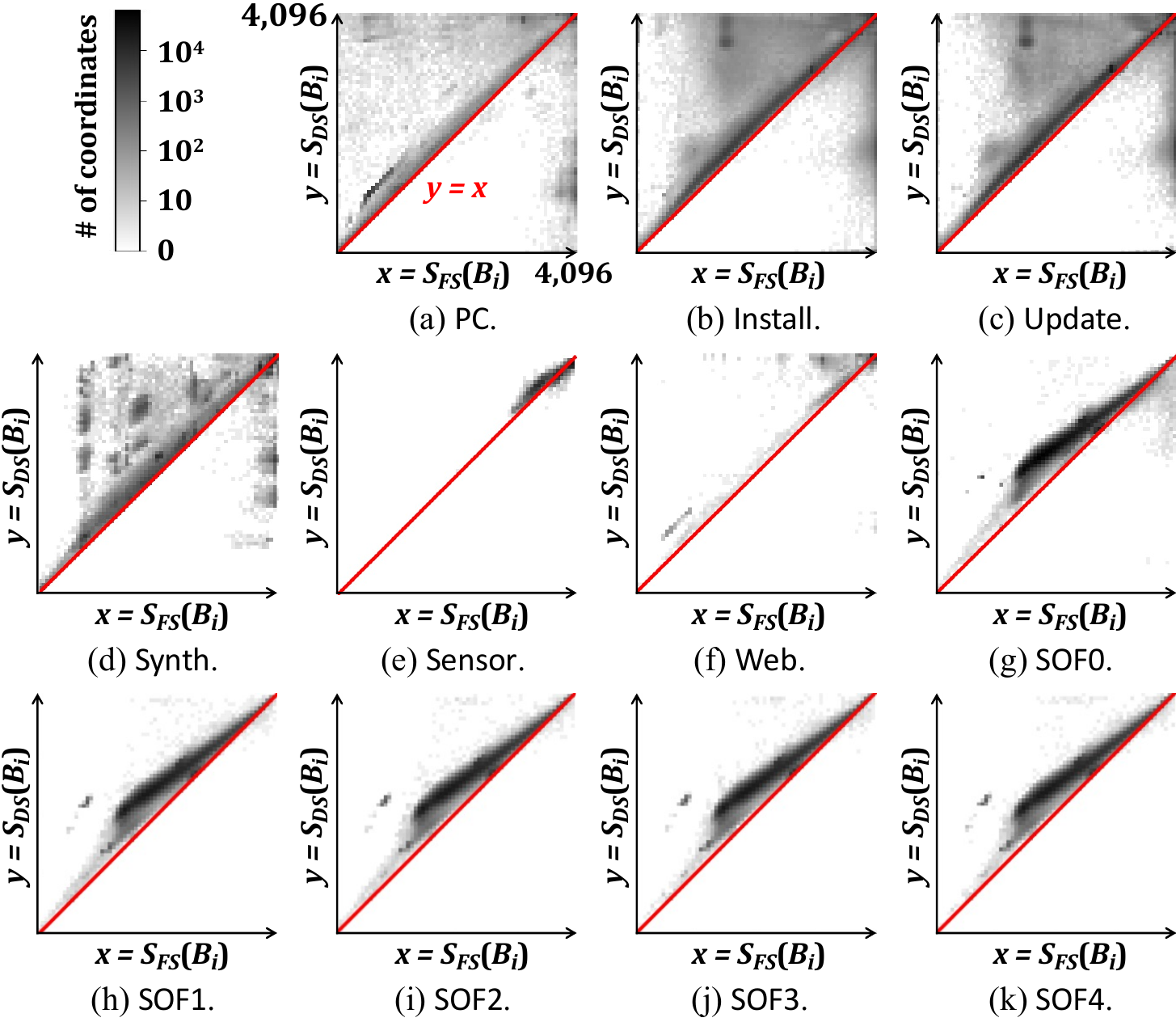}
    \vspace{-2em}
    \caption{Comparison of the reference-search pattern.}
    \label{fig:pattern}
    \vspace{-1.5em}
\end{figure}

\fig{\ref{fig:pattern}} plots coordinates of $x=$~$S_{FS}(B_i)$ and $y=$~$S_{DS}(B_i)$ for a block $B_i$ in each workload.
If $x = y$, Finesse and \prop exhibit the same delta-compression ratio (highlighted with a red line in \fig{\ref{fig:pattern}}), which implies that they select the same reference block. 
A coordinate ($S_{FS}(B_i)$, $S_{DS}(B_i)$) above (or below) the line means that \prop provides higher (or lower) data-reduction ratio for block $B_i$ than Finesse.

From \fig{\ref{fig:pattern}}, we make three observations.  
First, as expected, \prop provides higher data savings compared to Finesse for a large number of blocks under every workload. 
Second, despite the higher data savings of \prop over Finesse in general, there are also a non-trivial number of blocks for which Finesse selects better references, achieving higher data savings than \prop.  
Excluding the \sof{} workloads, Finesse selects higher-quality references compared to \prop for up to 11.8\% of the total blocks.
Third, \prop and Finesse show quite different patterns in reference search.
As shown in \fig{\ref{fig:pattern}}, the coordinates in $y > x$ region (\ie~where \prop outperforms Finesse) are close to the line $y = x$, and at the same time, many of them are scattered across a wide range of the region compared to the coordinates in $y < x$ region.
On the other hand, a majority of the coordinates in $y < x$ region (\ie~where Finesse outperforms \prop) tend to have a very large $y$ value (\eg~> 3,072).
These imply that, while Finesse is effective to find a reference highly similar to an input block, it also misses a number of blocks that \prop can find and use to improve the data-reduction efficiency. 

\vspace{-1em}
\subsection{Combination with Existing Techniques\label{subsec:combined}}
\vspace{-.5em}
Our second and third observations in \sect{\ref{subsec:pattern}} motivate us to combine \prop with existing techniques to maximize the data-reduction ratio.
We design a storage system that employs both Finesse and \prop.  
When the two techniques find different reference blocks for an incoming block, the system chooses the one that provides a higher data-reduction ratio.
Such an approach increases the memory and computation overheads for data sketching but would be desirable for a system where data reduction is paramount (\eg~backup systems).
We leave the study of efficiently combining \prop with existing techniques as future work.

\begin{figure}[h]
    \centering
    \vspace{-.5em}
    \includegraphics[width=\linewidth]{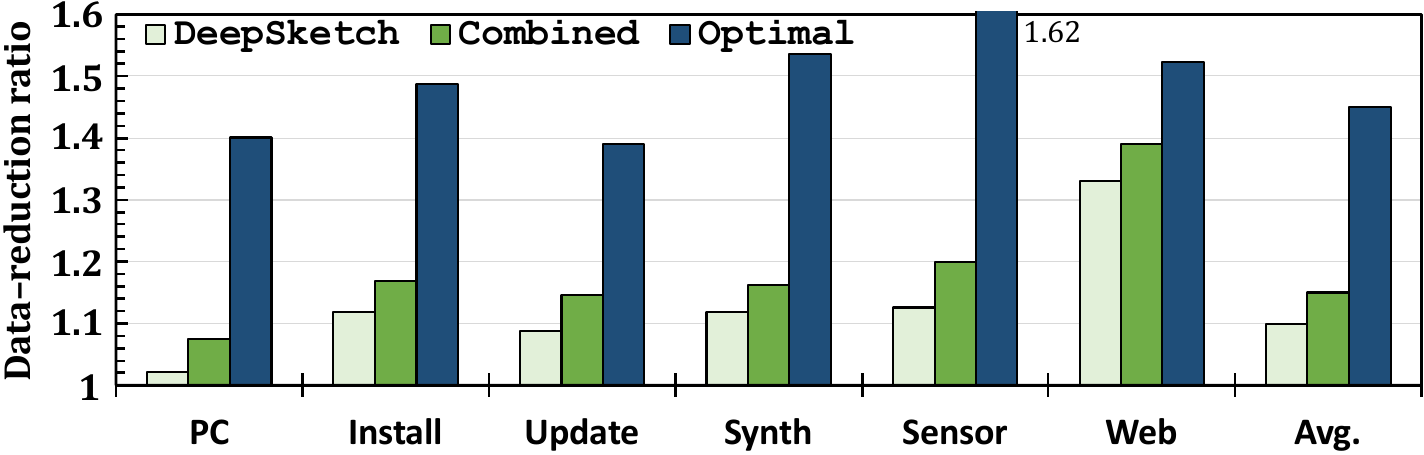}
    \vspace{-1.5em}
    \caption{Data-reduction improvement of a combination of \prop and Finesse.}
    \label{fig:combined}
    \vspace{-1em}
\end{figure}

\fig{\ref{fig:combined}} shows the combined approach's data-reduction benefits compared to when using either Finesse or \prop{} alone.\footnote{
We omit the results of the \sof{} workloads in \fig{\ref{fig:combined}} because there is no motivation to combine \prop{} with Finesse under such workloads for which Finesse provides negligible data reduction.} 
We also measure the \emph{optimal} data-reduction ratio (\ie~when every data block is delta-compressed with the best reference block found by brute-force search) for each workload to understand room for improvement after applying the combined approach.  
To emphasize the benefits of the combined approach over the standalone techniques, we normalize all the results in \fig{\ref{fig:combined}} to Finesse.

We observe that, as expected, the combined approach further improves the data-reduction ratio compared to the two standalone techniques under most workloads.
The combined approach achieves up to 38\% and 6.6\% (15\% and 4.8\% on average) data-reduction improvements over Finesse and \prop, respectively.
We also observe that the combined approach can reduce the gap in data-reduction ratios between the existing reference search techniques and the optimal.  
Although there is still large room for improvement (\ie~up to 35\% and 26\% on average) even after applying the combined approach, the combined approach reduces the gap by up to 81\% (\ie~62\% $\rightarrow$ 9.6\% under \web) and by 42\% on average. 
From our observations, we conclude that \prop can also be used as a useful method
to complement the weakness of existing post-deduplication delta-compression techniques.

\vspace{-1em}
\subsection{Impact of Training Data-Set Quality\label{subsec:tdata}}
\begin{figure}[!t]
    \centering
    \includegraphics[width=0.85\linewidth]{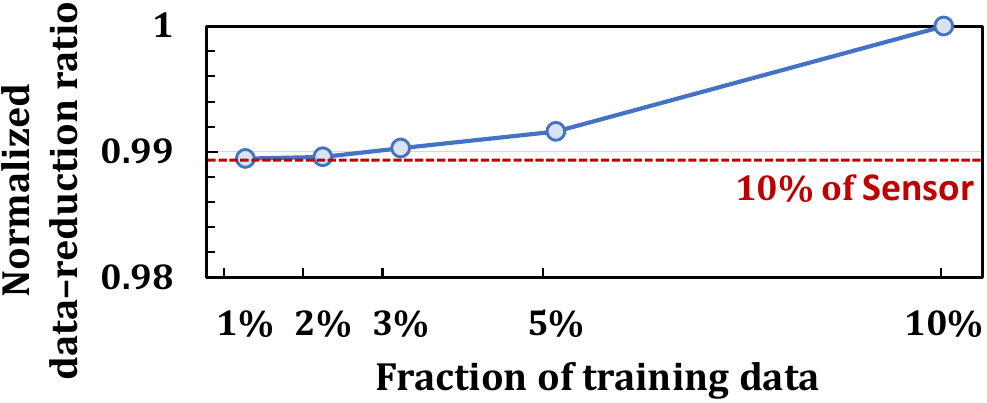}
    \vspace{-.5em}
    \caption{Effect of training data set on data-reduction ratio.}
    \vspace{-2em}
    \label{fig:tdata}
\end{figure}

\vspace{-.5em}
We evaluate the impact of the training data-set quality on the data-reduction ratio of \prop.
\fig{\ref{fig:tdata}} shows the average data-reduction ratio of \prop for the entire workloads listed in \tab{\ref{tab:workloads} when we use two different types of training data sets.
First, we evaluate how \prop's benefit changes when we train its DNN model using 1\%/2\%/3\%/5\%/10\% of the \emph{entire} data sets (the blue line in \fig{\ref{fig:tdata}}). 
Second, we measure \prop's benefit when we use 10\% of requests only from \sensor for DNN training (the dashed red line in \fig{\ref{fig:tdata}}).  
}
When we use $x$\% ($<$ 10\%) of each trace for training, we use the remaining $(100-x)$\% to evaluate the data-reduction ratio of \prop.
All values in \fig{\ref{fig:tdata}} are normalized to the data-reduction ratio obtained when using 10\% of the entire data sets for DNN training.

We make two observations from \fig{\ref{fig:tdata}}. 
First, while a larger training data set increases \prop's benefit, \prop can provide a fairly good data-reduction ratio even with a very small training data set. 
Using only 1\% of the traces for DNN training provides 98.9\% of the data-reduction ratio obtained when using 10\% of the traces.
Second, \prop can provide a high data-reduction ratio even when we use a training data set collected from a single trace. 
Compared to when we use 10\% of all traces for DNN training, using 10\% of only \sensor decreases the data-reduction ratio by less than 1\%.
Based on our observations, we conclude that it is possible to train an effective DNN model for \prop with a limited data set, while providing high adaptability for diverse input data sets.

\begin{figure}[h]
    \centering
    \vspace{-.5em}
    \includegraphics[width=\linewidth]{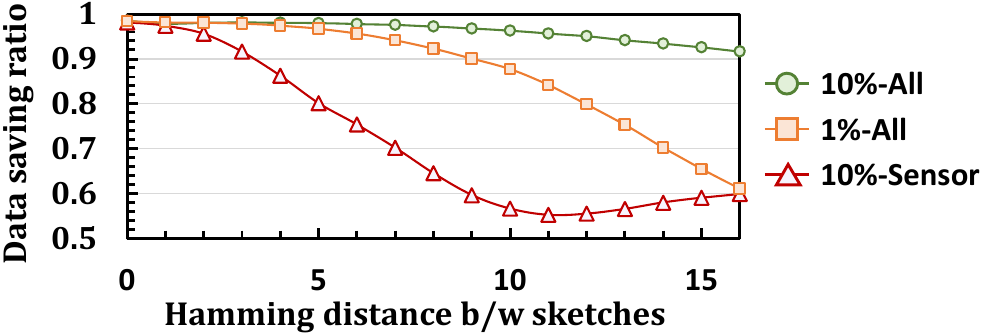}
    \vspace{-2em}
    \caption{Effect of training data set on sketch accuracy.}
    \vspace{-.5em}
    \label{fig:hdist}
\end{figure}

To study the detailed impact of the training data-set quality,
we analyze how the sketches generated by \prop change with different training data sets.
To this end, we measure the average \emph{data-saving ratio} (\ie~$1-$~\textit{Delta-Compressed Data Size} / \textit{Original Data Size}) of delta-compressed blocks depending on the Hamming distance between the sketches of the input and reference blocks (\ie~$\Delta(\mathbf{H}, \widehat{\mathbf{H}})$ in \sect{\ref{subsec:select}}.)
\fig{\ref{fig:hdist}} shows the relationship between the data-saving ratio and sketch Hamming distance for three different DNN models trained with 10\% of \sensor (\textsf{10\%-Sensor}) and 1\%/10\% of all traces (\textsf{1\%-All} and \textsf{10\%-All}).
In general, the higher the data-saving ratio at a low sketch Hamming distance, the more accurate the sketches generated by \prop.

We identify the following two findings from \fig{\ref{fig:hdist}}.
First, for all DNN models, \prop provides extremely high data saving (close to 1) when $\Delta(\mathbf{H}, \widehat{\mathbf{H}})\leq2$.
The result shows that all the three DNN models enable \prop to identify highly similar data blocks by generating almost identical sketches. 
It is due to the nature of the DNN-based learning-to-hash method: a DNN can be easily trained to yield the same hash values for the data with negligible differences.
Second, in \textsf{1\%-All} and \textsf{10\%-Sensor}, the data-reduction ratio degrades more significantly as the Hamming distance increases, compared to \textsf{10\%-All}.
It suggests that we can further improve the benefit of \prop by increasing the accuracy of the sketch generation with a better DNN model, \eg~using high-quality data sets and/or advanced model architectures.
In the current version of \prop, the ANN model compensates for such potential accuracy loss by finding sufficiently-good reference blocks with best efforts.

\vspace{-1em}
\subsection{Overhead Analysis\label{subsec:overhead}}
\begin{figure}[!h]
    \centering
    \vspace{-.5em}
    \includegraphics[width=\linewidth]{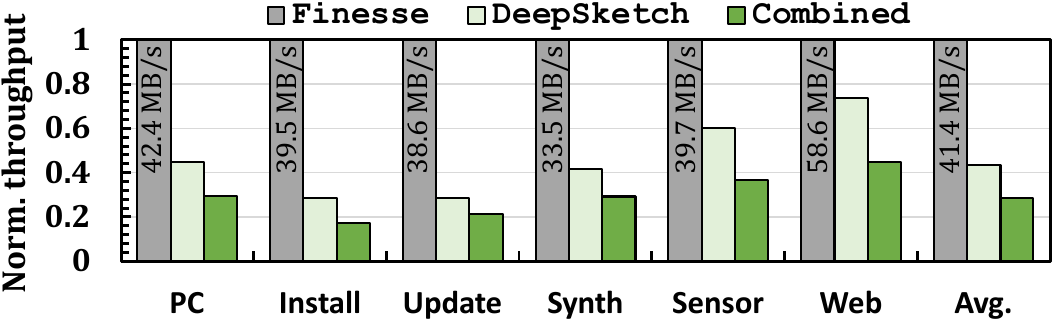}
    \vspace{-2em}
    \caption{Performance overhead of \prop.}
    \vspace{-1em}
    \label{fig:throughput}
\end{figure}

\vspace{-.5em}
\head{Performance Overhead}
\fig{\ref{fig:throughput}} shows the average throughput of \prop and the combined approach of \prop and Finesse under different workloads, normalized to Finesse.\footnote{Note that DNN training does \emph{not} affect \prop's throughput because it can be performed \emph{offline} as explained in \sect{\ref{sec:deepsketch}}. In our system described in \sect{\ref{subsec:methodology}}, DNN training (including \cluster) takes less than 4 hours with 300 epochs for our 1.6-GB training data set.}
\prop and combined approach provide up to 73.7\% and 44.9\% (44.6\% and 28.4\% on average across all workloads) of the average throughput of Finesse.
This non-trivial performance overhead is due to the inherent trade-off between the data-reduction ratio and throughput in post-deduplication delta compression; performing delta compression for more data blocks would increase the data-reduction ratio, but it comes at the cost of performance degradation since delta compression takes more time compared to lossless compression (\eg~in our current implementation, LZ4 takes 6.9~\usec per block on average, which is less than 10\% of the average execution time of Xdelta).

\begin{figure}[!h]
    \centering
    \vspace{-.5em}
    \includegraphics[width=\linewidth]{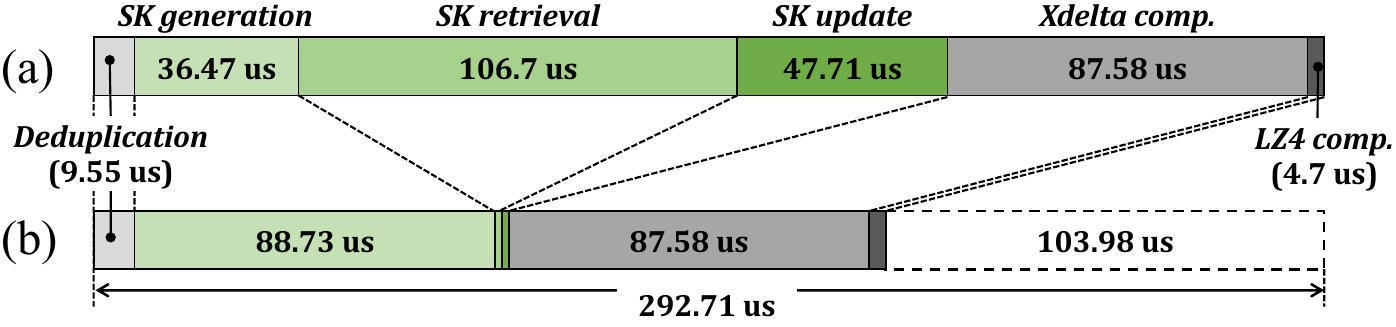}
    \vspace{-1.5em}
    \caption{Average latency for each data-reduction step in (a)~\prop and (b) Finesse.}
    \vspace{-1em}
    \label{fig:overhead}
\end{figure}

To better understand the performance overheads of \prop, we measure the average latency of each step per input data block during the post-deduplication delta-compression process.
\prop requires two modifications on existing techniques, 1) replacing the SF-based sketching engine with the DNN-based one and 2) using the ANN engine described in \sect{\ref{subsec:select}} as the SK store.
For fair comparison, we implement the SK store of Finesse using the unordered-map data structure that provides $O(1)$ time complexity for lookup.
\fig{\ref{fig:overhead}} visualizes the fraction of the average time spent for each step in the entire data-reduction process. 

As shown in \fig{\ref{fig:overhead}}, \prop and Finesse operate differently in only three steps of the entire data-reduction process: 1) sketch generation for an incoming block, 2) sketch retrieval from the SK store, and 3) sketch update to the SK store.
The other steps, including deduplication, Xdelta compression, and LZ4 compression, are performed in the same ways. 
Due to the simplicity of the hash network model network and GPU acceleration, \prop reduces the latency of sketch generation from 88.73 \usec to 36.47 \usec (by 58.9\%) over Finesse.
However, using the ANN engine significantly increases the latencies for sketch retrieval and update, leading to 55.1\% increase in the total average latency over Finesse. 

The performance overhead of \prop over Finesse is non-trivial, but it would not be a serious obstacle for its wide adoption due to two reasons.
First, we target a system where data reduction is critically important so that \prop's benefits outweigh its performance overheads.
Second, the performance overhead of \prop can be mitigated in several ways.
For example, if the data-reduction process is performed in \emph{background}, its negative performance impact could be relatively small.
We can also leverage the parallelism of multi-core CPUs to optimize software modules.
For example, the sketch update procedure can be performed in parallel with other modules. 
This hides the cost of updating sketches during the compression steps, thereby reducing the performance overhead by 45.8\% (\ie~103.98 \usec $\rightarrow$ 56.27 \usec).

\head{Memory Overhead}
Like existing post-deduplication delta-compression techniques~\cite{shilane-fast-2012, zhang-fast-2019}, \prop inevitably requires additional memory space for the sketch store. 
Despite the smaller sketch size of \prop compared to existing techniques~\cite{shilane-fast-2012, zhang-fast-2019} (128 bits vs. 192 bits), \prop's memory overhead might be unacceptable if it keeps track of the sketches of \emph{all} non-deduplicated data blocks.
For example, suppose that the data block size is 4 KiB, and 80\% of the stored data is unique (\ie~non-deduplicated). 
Then, the required memory space for the sketch store is about 0.3\% (0.8$\times$16/4,096) of the size of the stored data (e.g., around 100-GB memory space to work with 32-TB data).

However, the memory overhead would not be a significant obstacle \sj{to use \prop in practice for two reasons}.
First, the memory overhead for the sketch store is a common problem in \sj{all the sketch-based techniques}. 
Second, prior works demonstrate that a small fraction of data blocks are frequently used as the reference block for many input blocks~\cite{gupta-fast-2011, park-date-2017}.
Thus, keeping only most-frequently-used sketches in a limited-size sketch store (i.e., a with least-frequently-used (LFU) eviction policy) would provide sufficiently high compression efficiency.
We leave such further optimizations to mitigate \prop's memory overhead for future work. 

\vspace{-1em}
\section{Discussion\label{sec:discussion}}
\vspace{-.5em}

\head{Scalability to Larger Data Sets}
As shown in our evaluation, \prop provides high data-reduction ratios even for workloads that are not used in DNN training (\eg~SOF workloads), which implies the high generalizability of \prop.
Nevertheless, due to the limited amount of data sets publicly accessible, 
it is difficult to say whether \prop would be effective under \emph{any} given workload. 
For example, \prop may require a larger DNN model to provide sufficient benefits for large data sets (that we do not observe in this work), which would significantly increase the training overheads of \prop.
However, we believe that \prop would be able to work for larger data sets due to three reasons.
First, \prop’s DNN model has much smaller computation complexity than state-of-the-art DNN models, so there is significant room for \prop to use the larger DNN models. 
Second, the memory space required for training depends more on the size of the DNN model rather than the size of the training data set. 
Our current model is only a few hundreds of megabytes in size, which can be run on a single commonly-used GPU. 
Third, as explained in \sect{\ref{sec:deepsketch}}, DNN training can be performed offline in different machines with more computing/memory resources.  

\head{Cost-Effectiveness of \prop}
The current version of \prop requires a powerful GPU for DNN inference/training and thus introduces non-trivial performance and power overheads.
However, we believe that such overheads would not be a significant obstacle for the wide adoption of \prop due to two reasons.
First, as explained in \sect{\ref{sec:deepsketch}}, DNN training can be done in different machines (\eg~in cloud servers) without requiring frequent retraining, and multiple storage servers (that store similar types of data) can use the same DNN model, amortizing the training cost.
Second, there has been significant effort to develop high-performance and energy-efficient accelerators for both light-weight DNN inference (e.g.,~\cite{hadidi-iiswc-2019, qin-pr-2020, shafiee-isca-2016,boroumand-pact-2021}) and ANN search (e.g.,~\cite{kaul-isscc-2016, saikia-islped-2019, imani-icrc-2017}), which would greatly reduce the performance, power, and resource overheads of \prop.
\vspace{-1em}
\section{Related Work\label{sec:related}}
\vspace{-.5em}
To our knowledge, this work is the first to propose a learning-based data-sketching technique for accurate reference search in storage-level delta compression.
As we have already discussed state-of-the-art techniques closely related to \prop in Sections~\ref{sec:bg} and \ref{sec:motiv},
in this section, we briefly discuss other recent works on 1) storage-level data reduction and 2)~machine learning-based video/image compression.

\head{Storage-level Data Reduction}
The fundamental ideas of data-reduction techniques were proposed several decades ago.
Hence, their theoretical properties and limitations, in terms of data reduction, have been studied intensively.
Recent studies focus more on how to efficiently deploy them to various platforms
to achieve space savings with faster compression speeds, lower computation costs, and less energy consumption~\cite{zhang-fast-2019,chen-fast-2011,gupta-fast-2011,park-date-2017,lee-ieeetce-2011, yang-atc-2019}.
For example, SmartDedup~\cite{yang-atc-2019} proposes a low-cost deduplication technique for resource-constrained devices where computing resources as well as energy budget are seriously limited. 
Finesse~\cite{zhang-fast-2019} is a representative example of enhancing delta-compression speed without loss in data-reduction ratio by relaxing the complexity of sketch generation.
Some prior works~\cite{chen-fast-2011,gupta-fast-2011,lee-ieeetce-2011} present that deduplication and compression could be integrated in an SSD controller to improve storage lifetime as well as performance.

The key difference of this study from \rev{the above recent} works is that this work presents a new direction to \rev{improve \emph{data-reduction ratio}, the fundamental goal of a data-reduction technique.}
Our work analyzes the limitations of probabilistic and statistical approaches and shows that emerging deep-learning methods can be promising alternatives \om{to} and/or complements the traditional methods.

\head{Machine Learning for Video and Image Compression}
Several works attempt to improve video/image compression efficiency using machine learning~\cite{mentzer-neurips-2020,agustsson-iccv-2019, william-corr-2016, srivastava-icml-2015, chen-vcip-2017, chen-tcsvt-2020, srivastava-icml-2015}.
To enhance existing video compression algorithms, some leverage CNNs~\cite{chen-vcip-2017, chen-tcsvt-2020, srivastava-icml-2015} and others employ Long Short-Term Memory (LSTM) networks to learn video representation\om{s}~\cite{srivastava-icml-2015} and predict future frames~\cite{william-corr-2016}.  
Their common idea is to accurately predict pixel values of next video frames and store only delta\om{s} for reconstruction.
More recent studies use Generative Adversarial Networks (GAN) to generate part or all of \om{the} image \om{content} from a semantic label map~\cite{mentzer-neurips-2020,agustsson-iccv-2019}. 
They achieve space savings by storing only a smaller amount of preserved data and the label map in storage devices.

\prop is different from these studies in two aspects.  
First, unlike existing ML-based compression \om{methods} that target images and videos, \prop aims to compress binary data, which requires handling extremely high-dimensional data sets without any semantic information.
Second, ML-based compression \om{methods are} basically lossy compression algorithms, but our system is \om{a} lossless compression \om{system} that enables us to reconstruct original data without any data loss.
\vspace{-1em}
\section{Conclusion}
\vspace{-.5em}
We introduce \prop, the first learning-based reference search technique to improve the data-reduction efficiency of post-dedupli\-cation delta compression.
\prop uses the learning-to-hash method to overcome the limitations of existing techniques that miss a number of good reference candidates for delta compression of incoming data blocks.
We present a new \om{deep neural network} training method that enables \prop to efficiently learn delta-compression-aware data representation for unlabeled data sets with an extremely high dimensional space.
Using various real-world data sets, we experimentally demonstrate that \prop~\om{is} an efficient solution not only as a replacement \om{for} but also as a complement \om{to} existing reference search techniques, significantly reducing the data-reduction gap from the optimal.

\vspace{-1em}
\section*{Acknowledgements}
\vspace{-.5em}
We would like to thank our shepherd Chao Tian and anonymous reviewers for their feedback and comments.
We thank the SAFARI Research Group members for feedback and the stimulating intellectual environment they provide.
We thank our industrial partners, especially Google, Huawei, Intel, Microsoft, and VMware, for their generous donations. 
This work was in part supported by \om{the Semiconductor Research Corporation,} SNU-SK Hynix Solution Research Center (S3RC)\rev{,} and the National Research Foundation (NRF) of Korea (NRF-2018R1A5A1060031, NRF-2020R1A6A3A03040573).
(\emph{Co-corresponding Authors: Sungjin Lee and Onur Mutlu})
\vspace{-.5em}

\bibliographystyle{plain}
\bibliography{references}

\balance{}

\end{document}